\documentclass[journal]{IEEEtran} %Ë«À¸
\usepackage{textcomp}
\usepackage{booktabs}
\usepackage{colortbl}
\usepackage[table]{xcolor}
\usepackage{graphicx}
\usepackage{subfigure}
\usepackage{multirow}
\usepackage{longtable}
\usepackage{algorithm}
\usepackage{algorithmic}
\usepackage{booktabs}
\usepackage{stfloats}
\usepackage{caption2}
\usepackage{amssymb,amsmath}
\usepackage{epstopdf}
\usepackage{color}
\usepackage{rotating}
\usepackage{multirow}
\usepackage{threeparttable}
\usepackage{bbding}
\usepackage{tabularx}
\usepackage{booktabs}
\usepackage{cite}
\usepackage[colorlinks,linkcolor=red,anchorcolor=blue,citecolor=blue]{hyperref}

\usepackage{orcidlink}

\definecolor{rrr}{rgb}{1    0	0.0392}
\definecolor{myred}{rgb}{0.7059    0	0.0392}					% revised contents for reviewer 1
\definecolor{myblue}{rgb}{0    0	0.9392}					% revised contents for reviewer 1
\ifCLASSINFOpdf
\else
\fi
\begin{document}
\title{GLCONet: Learning Multi-source Perception Representation for Camouflaged Object Detection}
%\title{Co-Optimization: Learning Multi-source Perception Representation for Camouflaged Object Detection}
\author{
        Yanguang~Sun$^{\orcidlink{0009-0006-3765-6646}}$, 
        Hanyu~Xuan,
	Jian~Yang, %\emph{Member,IEEE},
        Lei~Luo %\textsuperscript{\Envelope}
%        Lebing~Zhang,
%        Yue~Li
%        and Xingming~Sun, \emph{Senior Member, IEEE}
\thanks{This work was supported by the National Natural Science Foundation of China (Grant No. 62276135, 61806094, and 62176124), %Natural Science Research Project of Colleges and Universities in Anhui Province (Grant No. xxxxxxx). 
(\textit{Corresponding author: Lei Luo,} e-mail: luoleipitt@gmail.com)
	
Y. Sun, L. Luo, and J. Yang are with the PCA Laboratory, Key Laboratory of Intelligent Perception and Systems for High-Dimensional Information of Ministry of Education, School of Computer Science and Engineering, Nanjing University of Science and Technology, Nanjing, China. (e-mail: Sunyg@njust.edu.cn; luoleipitt@gmail.com; csjyang@mail.njust.edu.cn).

H. Xuan is with the School of Big Data and Statistics, Anhui University, Hefei, 230601, China. (22176@ahu.edu.cn)

%J. Yang is with the College of Computer Science and Engineering, Nankai University, Tianjin, 300000, China. (csjyang@nankai.edu.cn)
}
}

\markboth{Journal of \LaTeX\ Class Files}%
{Shell \MakeLowercase{\textit{et al.}}: Bare Demo of IEEEtran.cls for IEEE Journals}

\maketitle
\begin{abstract}
%Recently, biological perception-based methods have shown impressive performance in camouflaged object detection (COD). However, most of them can only exploit limited spatial local information from convolutional operations to optimize initial multi-level features and ignore the long-range dependencies that can help the model build global connections and relationships between feature pixels from different scale-spaces, leading to inaccurate predictions. In this work, we start from a different view and propose a novel Global-Local Collaborative Optimization Network, called GLCONet. Technically, we first design a collaborative optimization strategy (COS) to simultaneously model the local spatial information and global long-range relationships, which can provide features with abundant discriminative information to boost the accuracy in detecting camouflaged objects. Furthermore, we introduce an adjacent reverse decoder (ARD) that contains cross-layer aggregation and reverse optimization to integrate complementary information from different levels for generating high-quality representations. Extensive experiments demonstrate that the proposed GLCONet method with different backbones, effectively activates significant pixels, outperforming 19 state-of-the-art (SOTA) methods on three public COD datasets.

Recently, biological perception has been a powerful tool for handling the camouflaged object detection (COD) task. However, most existing methods are heavily dependent on the local spatial information of diverse scales from convolutional operations to optimize initial features. A commonly neglected point in these methods is the long-range dependencies between feature pixels from different scale spaces that can help the model build a global structure of the object, inducing a more precise image representation. In this paper, we propose a novel Global-Local Collaborative Optimization Network, called GLCONet. Technically, we first design a collaborative optimization strategy from the perspective of multi-source perception to simultaneously model the local details and global long-range relationships, which can provide features with abundant discriminative information to boost the accuracy in detecting camouflaged objects. Furthermore, we introduce an adjacent reverse decoder that contains cross-layer aggregation and reverse optimization to integrate complementary information from different levels for generating high-quality representations. Extensive experiments demonstrate that the proposed GLCONet method with different backbones can effectively activate potentially significant pixels in an image, outperforming twenty state-of-the-art methods on three public COD datasets. The source code is available at: \href{https://github.com/CSYSI/GLCONet}{\color{blue} https://github.com/CSYSI/GLCONet}.

\end{abstract}
\IEEEpeerreviewmaketitle

\section{INTRODUCTION}
\IEEEPARstart{C}{amouflage} is an innate protective mechanism in some organisms ($e.g.$, chameleons, stingrays, and caterpillars) that is evolved to avoid danger from predators. Camouflaged object detection (COD) is a very challenging task that aims to detect concealed objects that have high similarity to their surroundings, and has been widely used in many applications, including medical image segmentation \cite{MIS}, crack inspection \cite{cd}, $etc.$ How to identify subtle differences between objects and backgrounds are crucial for the accurate COD task.

Early COD methods \cite{E_1,E_2,E_3} focus on hand-crafted features or expert knowledge to predict camouflaged objects, but their accuracy is unsatisfactory. Subsequently, with the development of deep learning and the open source of some large-scale datasets ($e.g.$, CAMO \cite{CAMO} and COD10K \cite{COD10K}), numerous COD methods based on deep learning begin to emerge, among which bio-inspired methods \cite{COD10K,BSANet,Bi-RRNet,TNNLS1,DCCFNet} is one of the most representative strategies. These bio-inspired COD methods achieve great performance and integrate features from different receptive fields to increase local spatial information in initial multi-level features through a series of well-designed convolutional modules. For example, the Modified RF is proposed in the search module of SINet \cite{COD10K}, the multi-scale feature extractor (RMFE) is designed in BSANet \cite{BSANet}, the multi-scale feature aggregation module is developed in FAPNet \cite{FAPNet}, the multi-scale scene perception module (MSPM) in Bi-RRNet \cite{Bi-RRNet}. In addition, some other COD methods \cite{SegMaR,ZoomNet,FDER} have utilized receptive field block (RFB) \cite{RFB}, atrous spatial pyramid pooling (ASPP) \cite{ASPP}, or dense atrous spatial pyramid pooling (DenseASPP) \cite{DenseASPP} for enhancing the performance of initial features by integrating multi-receptive field information. 

\begin{figure*}[]
	\centering\includegraphics[width=0.98\textwidth,height=4.3cm]{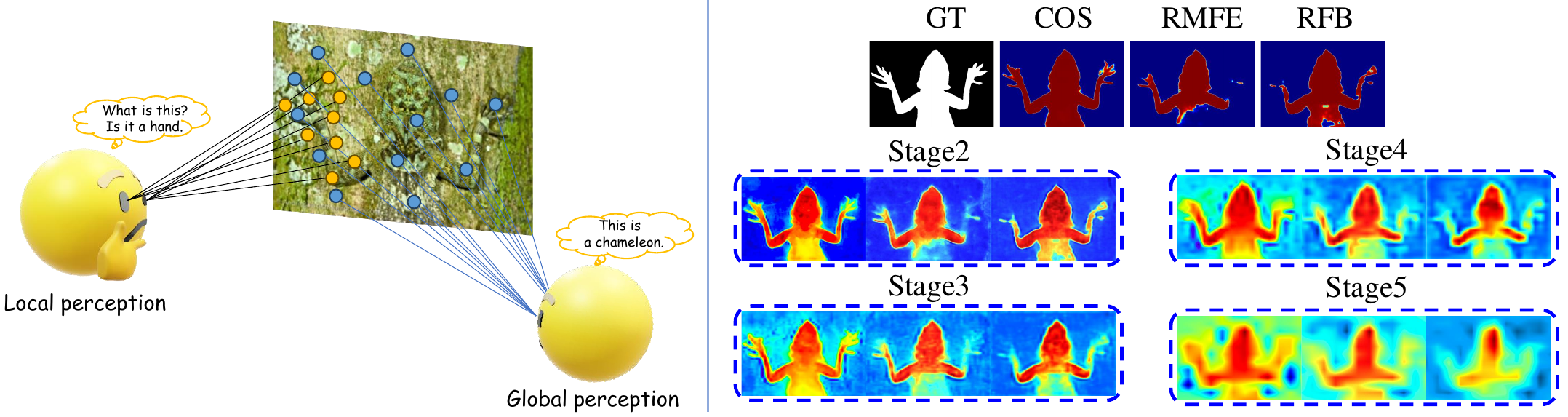}
	\captionsetup{font={small}}
	\caption{Local and global perception visual comparisons. Visual results between our COS and existing well-designed convolutional modules. ``Stage 2'' - ``Stage 5'' denote multi-level features from different modules. ``GT'' presents the ground truth. }
	\label{Fig.1}
\end{figure*}

However, the features optimized through the above methods often exhibit a local attribute, due to local connectivity and translation equivariance in the structure of the convolutional neural network (CNN) \cite{ViT,VST}. As shown on the left side of Fig. \ref{Fig.1}, features with a local perspective contain rich spatial details, which usually allows only partial identification of the object. When the object pixels are far apart, it becomes extremely difficult to achieve whole segmentation. Although the receptive field can be enlarged by increasing the size and filling rate of the convolutional kernel, such receptive fields are still localized and are unable to obtain global relationships in a real sense, which is not conducive to the segmentation of camouflaged objects with diverse sizes and types. As depicted on the right side of Fig. \ref{Fig.1}, we visualize the prediction feature maps of RFB \cite{RFB} and RMFE \cite{BSANet} across four stages, ranging from low-level to high-level features. Specifically, when the range of camouflaged objects is relatively large, the RFB \cite{RFB} and RMFE \cite{BSANet} that only contain convolutional operation will cause the incomplete problem of objects in the feature of different stages. Many studies \cite{yang2016nuclear,luo2016robust, luo2016tree,li2016double,Swin} have shown that capturing the global structure of images in modeling can obviously improve the model performance. Considering that global perception is important for image understanding, some COD methods \cite{FSPNet,VST,EVP,FPNet} use a transformer as an encoder to obtain long-range relationships, such as vanilla vision transformer (ViT) in FSPNet \cite{FSPNet}, pyramid vision transformer (PVT) in FPNet \cite{FPNet}, and SegFormer in EVP \cite{EVP}. However, global information should not be considered only in the encoder since COD belongs to the dense prediction task adopting typical encoder-decoder structure, each component of which is crucial for object detection and segmentation. For better segmenting camouflaged objects from complex surroundings, diverse perception information ($i.e.$, local details and global relationships) is also needed in the decoder stage to optimize initial features from the encoder to generate high-quality prediction maps. 

In this paper, we propose a novel collaborative optimization strategy (COS) that is capable of simultaneously modeling long-range relationships and local details in the COD task. As depicted in Fig. \ref{Fig.1}, our COS can effectively infer camouflage objects from complex backgrounds at each stage, benefiting from the collaborative optimization between global and local perception information. Technically, we design a global perception module (GPM) to capture long-range relationships through the multi-scale transformer block (MTB) that contains a multi-scale self-attention and a multi-scale feed-forward network. The advantage of using a multi-scale strategy is that it can enrich and diversify features \cite{ZoomNet}, providing rich contextual information for camouflaged object segmentation. Next, a local refinement module (LRM) extracts spatial details from different receptive fields in a local manner by utilizing a well-designed progressive convolution block (PCB). Furthermore, %we aggregate these global-local information in the group-wise hybrid interaction module (GHIM) to enhance the feature expression ability. 
we aggregate these multi-source information ($i.e.$, local details and global relationships) in the group-wise hybrid interaction module (GHIM) to enhance the feature expression ability. In addition, we propose an adjacent reverse decoder (ARD) with cross-layer aggregation and reverse optimization to mitigate problems
related to semantic dilution and weak aggregation in the traditional feature pyramid network decoder for accurate COD tasks. Extensive experimental results on three COD benchmarks show that our GLCONet method has remarkable advantages in terms of multiple evaluation metrics when compared to twenty state-of-the-art (SOTA) COD methods under different backbone networks. 

In a word, our main contributions are summarized as:
\begin{itemize}
	\item We propose a Global-Local Collaborative Optimization Network that generates high-quality features by activating potentially significant pixels from the perspective of multi-source perception for better detecting objects. 
	
	\item We design a collaborative optimization strategy that is capable of modeling and aggregating global and local information through the joint operation of GPM, LRM, and GHIM.
	
	\item We develop an adjacent reverse decoder that integrates the complementary information from different level features by employing cross-layer aggregation and reverse optimization.
\end{itemize}

The remainder of this paper is organized as follows. Section II gives the related research works, while Section III presents the details of the proposed GLCONet method. Experimental results are shown and discussed in Section IV. Finally, the concluding remarks are shown in Section V.

\section{Related Work}

\subsection{Camouflaged Object Detection}
Early traditional COD methods \cite{E_1,E_2,E_3} typically rely on hand-crafted features or low-level visual priors ($e.g.$, color, texture, and intensity) to detect camouflaged objects. However, these features or priors may not fully obtain the complex structures and content within the image. Especially, camouflaged objects in natural images often have an extremely high similarity with the background, leading to unsatisfactory segmentation results with traditional strategies.

%Early COD tasks \cite{E_1,E_2,E_3} mainly used low-level priors or hand-crafted features ($e.g.$, color, shape and intensity) to obtain camouflaged objects. However, in the real world, camouflaged or concealed targets have only slight differences from their surroundings, making early COD methods unsatisfactory. 

%To address this difficult problem, 
Later, thanks to the availability of large-scale datasets ($i.e.$, CAMO \cite{CAMO}, COD10K \cite{COD10K}, and NC4K \cite{NC4K}), numerous deep learning-based COD methods \cite{SegMaR,FSNet,TNNLS2,FSEL} have been proposed to address this difficult segmentation problem. One of the most representative methods \cite{COD10K,BSANet,FAPNet,Bi-RRNet} is biologically inspired. These methods aim to aggregate multi-scale features captured by diverse convolutional operations to gradually distinguish between objects and backgrounds. Specifically, referring to the human visual perception system, SINet \cite{COD10K} proposed a search module and identification module to locate and segment camouflaged objects. BSANet \cite{BSANet} designed a residual multi-scale feature extractor to obtain multi-scale features for better understanding image content in COD tasks. FAPNet \cite{FAPNet} developed a multi-scale feature aggregation module to adaptively extract multi-scale learning from different levels for dealing with scale variations. Bi-RRNet \cite{Bi-RRNet} exploited a multi-scale scene perception module to mitigate the significant object appearance variations by adjusting the multi-scale feature from each layer. PCPNet \cite{TNNLS1} utilized the vital component generation module to obtain a set of compact features rich in low-high level information across diverse subspaces. The above approaches \cite{COD10K,BSANet,FAPNet,Bi-RRNet,TNNLS1} have achieved better performance compared to traditional methods by integrating spatial information that contains different receptive fields.
%Other existing COD methods have also achieved great performance through employing different strategies ($e.g.$, multi-stage \cite{SegMaR}, joint training \cite{USCO} and loss function \cite{ZoomNet}. 
However, convolutional operations tend to be limited in the receptive field, and often the extracted cues are localized making it difficult to model global relationships between all pixels. Especially when the model's receptive field is limited to a local perspective and lacks a global view, the predicted camouflage object may be incomplete as a whole. Different from the above methods, we propose a collaborative optimization strategy that adopts a dual branch approach ($i.e.$, GPM, LRM, and GHIM) to simultaneously capture local spatial details and global semantic relationships for jointly optimizing initial features.  Furthermore, we utilize an adjacent reverse decoder to interact and aggregate diverse information from multiple layers via cross-layer aggregation and reverse optimization.

\subsection{Vision Transformer}
Transformer is first used in the field of natural language processing \cite{T} and has achieved remarkable success. Later, thanks to the efficiency of self-attention, Transformer has been widely utilized in numerous computer vision tasks, including image generation \cite{IG}, image captioning \cite{ASET}, semantic segmentation \cite{CoT}, image deblurring \cite{IM}, $etc$. For example, ViT \cite{ViT} performed the image classification task by modeling long-range dependencies from the sequence of image patches. Restormer \cite{Restormer} designed the dconv head transposed attention module to capture global context and gated-dconv feed-forward network to perform controlled feature transformation for the image restoration task. In addition, other Transformer models have been successful in computer vision, such as HAT \cite{HAT}, EVT \cite{EVT}, BiFormer \cite{BiFormer}, Pyramid ViT \cite{PVT}, CrossFormer \cite{Cvt}, and among others. 

Recently, Transformers have demonstrated excellent segmentation performance in COD tasks. Typically, FSPNet \cite{FSPNet} utilized a vanilla vision transformer \cite{ViT} as the encoder to model the global context of camouflaged objects. EVP \cite{EVP} adopted recent SegFormer \cite{segformer} to take the features from the frozen patch embedding and the input’s high-frequency components as prompting for low-level structure segmentation. In addition, FPNet \cite{FPNet} and FSNet \cite{FSNet} respectively apply PVT \cite{PVT} and Swin Transformer \cite{Swin} as encoders to obtain representations with global attributes from input images for detecting camouflaged objects. Although existing Transformer encoder-based methods have further improved performance in COD tasks, we believe that global exploration beyond the encoder is also crucial for accurate segmentation tasks. Considering these, we design a multi-scale transformer block that models long-range relationships between all pixels in multi-scale spaces to enhance the global perception ability of input initial features for better inferring camouflaged objects with diverse types and variable sizes.
\begin{figure*}
	\centering\includegraphics[width=0.84\textwidth,height=8.5cm]{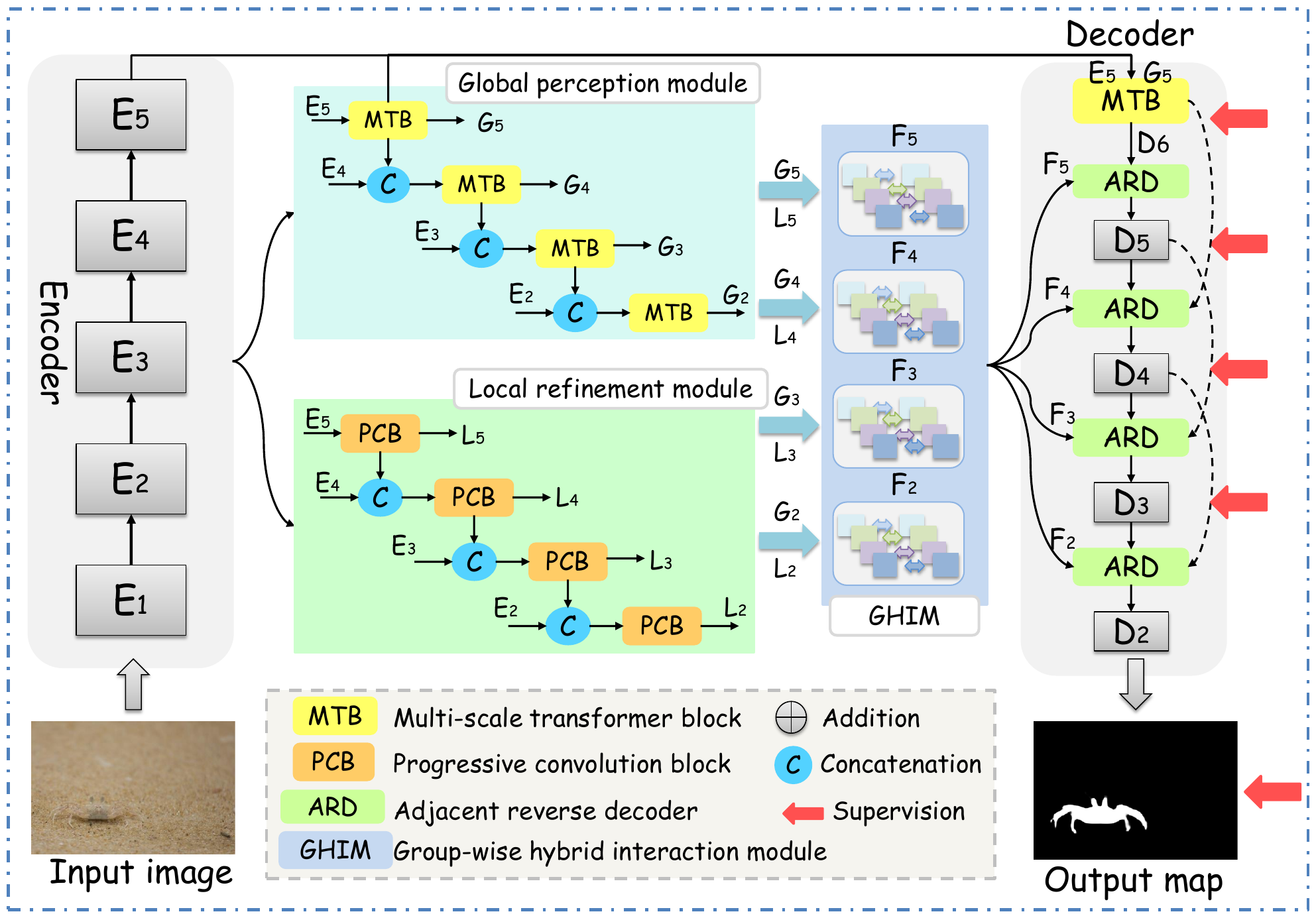}
	\captionsetup{font={small}}
	\caption{Overall architecture of our GLCONet method. We use ResNet-50/Swin Transformer/PVT as the encoder and propose a collaborative optimization strategy (COS) that contains a global perception module (GPM), a local refinement module (LRM) and a group-wise hybrid interaction module (GHIM) to simultaneously model long-range dependencies and local details. In addition, we design an adjacent reverse decoder (ARD) to integrate the complementary information with different layers through cross-layer aggregation and reverse optimization.}
	\label{Fig.2}		
\end{figure*}
\begin{figure*}
	\centering\includegraphics[width=0.90\textwidth,height=5cm]{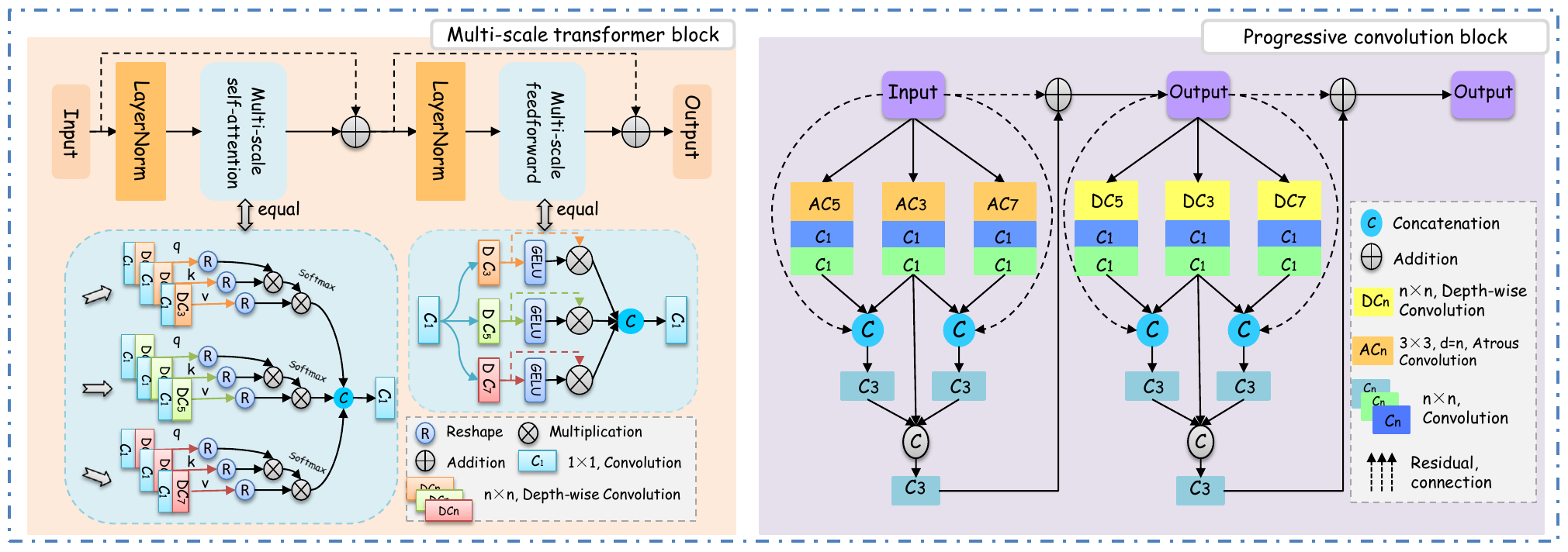}
	\captionsetup{font={small}}
	\caption{Details of the multi-scale transformer block (MTB) and the progressive convolution block (PCB).}
	\label{Fig.3}		
\end{figure*}

\section{Proposed GLCONet Method}
To alleviate the limitation that the existing feature optimization modules ignore long-range relationships between feature pixels in different scale spaces, we propose a new collaborative optimization strategy, dubbed COS. The COS adopts two different mechanisms to simultaneously capture long-range dependencies and local details, and aggregates global-local information using a group interaction manner to improve the discriminative ability of features for all pixels. In addition, we design an adjacent reverse decoder (ARD) to effectively decode the multi-source information from the optimized features to generate high-quality representations for accurate segmentation of camouflaged objects.

In the following sections, we first give an overview of our GLCONet, then introduce the details of our COS and ARD, and the loss function used for training, respectively.

\begin{table}[]
 \centering
 \caption{Summary and implication of symbols}.
%\begin{tabular}{lc}{\textwidth}{p{0.08\textwidth}X}
\begin{tabularx}{0.48\textwidth}{p{0.08\textwidth}X}
\toprule
\textbf{Symbols} & \textbf{Corresponding implications} \\ 
$\left \{ E_{i} \right \} _{i=1}^{5} $     &  initial multi-level features from the encoder     \\ 
$\left \{ F_{i} \right \} _{i=2}^{5}$     & multi-level features optimized through the COS      \\ 
$\left \{ D_{i} \right \} _{i=2}^{5}$     &  feature maps without activating from the ARD    \\
$D_6$     &   a coarse global feature map from the MTB    \\ 
$E_{i}^{128}$     & the input initial feature with 128 channels   \\ 
$\widetilde{E_{i}}$ & the optimized feature through a layer normalization \\
$FN_{ni}$  & intermediate process features from the multi-scale FFN \\
$\widehat{E_{di}}, P_{3i}^{k}$ & local spatial features from the first stage of the PCB \\
$Y_{3i}^{k}, \widehat{L_{ki}^{1}}$ & local spatial features from the second stage of the PCB \\
$G_{i}^{1},G_{i}^{2}$ &  global features of the two stages in the MTB \\
$L_{i}^{1},L_{i}^{2}$ & final local features of the two stages in the PCB \\
$G_i, L_i$ & global semantic features, local spatial features \\
$ G_{i}^{m}$ &  global features after splitting in the GHIM  \\
$L_{i}^{m}$ &  local features after splitting in the GHIM\\
$F_{i}^{m}$ &  global-local features after fusing in the GHIM \\
$Q_{ni}, K_{ni}, V_{ni}$ & different scale $query$, $key$, and $value$ \\
$A_{ni}$ & transpose-attention maps with different scale \\
$\mathcal{LN}(\cdot)$      &   layer normalization    \\ 
$C_n$      &  $n \times n$ convolution     \\ 
$DC_n$ & $n\times n$ depth-wise convolution \\ 
$AC_d$ & $3 \times 3$ atrous convolution with filling rate of $d$ \\
$GC_3$ & $3 \times 3$ gated convolution \\
$\Phi(\cdot), +$ & concatenation, element-wise addition  \\
$\eth\left|\cdot \right|, \odot$ & the GELU non-linearity, element-wise multiplication \\
$\varrho(\cdot)$ & multiple convolution operations \\
$\mho(\cdot)$ & pixel-shuffle upsampling operation \\
$\Lambda, F_{i}^{r}$ & reversed attention map, reverse feature \\
\toprule
\end{tabularx}
\label{TABLE1}
\end{table}

\subsection{Overview}
Fig. \ref{Fig.2} illustrates the overall architecture of our GLCONet, where the main structure is based on an encoder-decoder. Specifically, input an image ${I}\in\mathbb{R}^{H\times W\times C}$, the basic encoder ($i.e.$, ResNet-50 \cite{ResNet}/Swin Transformer \cite{Swin}/ Pyramid Vision Transformer \cite{PVT}) generates initial features $\left \{ E_{i} \right \} _{i=1}^{5} $ with the resolution of $\frac{W}{2^{i}} \times \frac{H}{2^{i}}$. Note that due to the large resolution and background noise of the first layer feature $E_1$, we discard the feature $E_1$ and utilize four features $\left \{ E_{i} \right \} _{i=2}^{5}$ for the COD task. Subsequently, Our COS generates discriminative features $\left \{ F_{i} \right \} _{i=2}^{5}$ with 128-channel. Additionally, the features $E_5$ and $G_5$ are input into a multi-scale transformer block (MTB) to predict a coarse feature map $D_6$. Finally, features $\left \{ F_{i} \right \} _{i=2}^{5}$ are fed into the ARD to produce feature maps without activating $\left \{ D_{i} \right \} _{i=2}^{5}$.  Through the collaborative optimization of multiple components, our GLCONet method achieves significant performance even in the face of challenging objects. To enhance the readability of the paper, we provide explanations for all symbols and their corresponding implications in Table \ref{TABLE1}.

\subsection{Collaborative Optimization Strategy}
Biological-based models have shown powerful performance in COD tasks, which apply in numerous convolutions to capture local multi-receptive field information. However, the inherent structure of convolutions often makes them obtain restricted receptive fields and difficult to capture global relationships of all pixels, resulting in inaccurate predictions. For that, we propose the COS that contains three components, that is, a global perception module (GPM), a local refinement module (LRM) and a group-wise hybrid interaction module (GHIM). The first two modules explore global and local perception representations through different structures, while the latter is utilized to integrate the global-local information.

\subsubsection{Global perception module}
Technically, we design a global perception module (GPM), which utilizes the multi-scale transformer block (MTB) to obtain the relationship of all pixels from a global angle. As depicted in Fig. \ref{Fig.3}, GPM structure is top-down to increase information flow, and its internal component mainly contains four MTBs. 

MTB is inspired by Restormer \cite{Restormer}, which has achieved great success in the image super-resolution (SR). Considering the differences between SR and COD tasks, camouflaged objects in images often have variable shapes and unfixed scales, requiring massive contextual information for understanding image content. Therefore, we propose a multi-scale self-attention and a multi-scale feed-forward network into MTB. The purpose of mapping features to different scale spaces is to enrich input features and increase their diversity, providing more diverse contexts for identifying camouflaged objects. Specifically, for an input feature $E_{i}^{128}$ with 128-channel after dimensionality reduction, a layer normalization $\mathcal{LN}$ is first used to produce a tensor $\widetilde{E_{i}}$, that is, $\widetilde{E_{i}}$=$\mathcal{LN}$ ($E_{i}^{128}$). Next, our MTB generates different scale $query$ ($Q_{ni}$=$C_{1}^{Q}{DC_{n}^{Q}} \widetilde{E_{i}}$), $key$ ($K_{ni}$=$C_{1}^{K}{DC_{n}^{K}}\widetilde{E_{i}}$) and $value$ ($V_{ni}$=$C_{1}^{V}{DC_{n}^{V}} \widetilde{E_{i}}$) projections through adopting 1$\times$1 point-wise convolution ($C_1^{(\cdot)}$) followed $n\times n$ depth-wise convolution ($DC_n^{(\cdot)}$). Based on the experience of the RFB \cite{RFB} method, the $n$ is set to 3, 5, and 7, respectively. Subsequently, we independently reshape $query$ ($\widehat{Q_{ni}}\in\mathbb{R}^{C\times HW}$) and $key$ ($\widehat{K_{ni}}\in\mathbb{R}^{HW\times C}$) projections such that their dot-product interaction produces three transpose-attention maps ($A_{ni}\in\mathbb{R}^{C\times C}$). Finally, we perform attention map activation and concatenate three attention features with different scales to generate the first stage feature $G_{i}^{1}$, which can be formulated as follows:
\begin{equation}
\begin{split}
	&G_{i}^{1}=C_{1}\Phi(A_{3i}\widehat{V_{3i}},A_{5i}\widehat{V_{5i}},A_{7i}\widehat{V_{7i}})+ E_{i}^{128}, \\
	&A_{ni} = Softmax\left (\widehat{Q_{ni}}\widehat{K_{ni}} / \sqrt{d_{ni}} \right), 
\end{split}
\end{equation}
where $\Phi(\cdot,\cdot,\cdot)$ denotes the concatenation. $C_1$ is 1 $\times$1 convolution. $\sqrt{d_{ni}}$ is a scaling factor that keeps the result from being too large or too small. $\widehat{V_{ni}}$ represents the reshaped $V_{ni}$ and its size is $\mathbb{R}^{C\times HW}$. ``$+$'' presents the element-wise addition.

To improve the expressive ability of the feature, we introduce a multi-scale operation in the feed-forward network (FFN). Similarly, a layer normalization $\mathcal{LN}$ is first adopted to generate a tensor $\widetilde{G_{i}^{1}}$, and then we exploit multiple depth-wise convolutions to restructure features and use a gating mechanism and the GELU non-linearity function to further enhance performance. Ultimately, the second stage feature $G_{i}^{2}$ with abundant global contexts is generated via concatenating features at different scales. $G_{i}^{2}$ is formulated as:
\begin{equation}
\begin{split}
	&G_{i}^{2} = C_1\Phi (FN_{3i},FN_{5i},FN_{7i})+ G_{i}^{1},\\
	&FN_{ni}=\eth \left |  (C_1DC_n(\widetilde{G_{i}^{1}}) )\right |\odot C_1DC_n(\widetilde{G_{i}^{1}}), 
	\end{split} 
\end{equation}
where $\eth\left|\cdot \right|$ is the GELU non-linearity function. ``$\odot$'' denotes the element-wise multiplication.
Finally, we introduce residual connections to generate the final feature $G_i$, that is, $G_{i}$=$C_{3}\Phi(G_{i}^{2},E_{i}^{128})+E_{i}^{128}$.

\subsubsection{Local refinement module}
Convolutional operations with local receptive fields, shared weights, and spatial sampling can be forced to capture the local structure. Therefore, we propose a local refinement module (LRM) to increase the spatial local information in initial features. 

Unlike RFB \cite{RFB} and RMFE \cite{BSANet} that directly combine all features, our LRM captures local spatial details by utilizing the progressive convolution block (PCB) of two stages, which aims to obtain multi-source local information from different operations. As shown in Fig. \ref{Fig.3}, the structure and input of the LRM are the same as the GPM. Specifically, in the first stage, the feature $E_{i}^{128}$ is input in the PCB, and then we adopt multiple 3$\times$3, atrous convolution ($AC_{d}$) with filling rate of $d$ and followed two 1$\times$1 convolutions ($C_1$) to generate local feature $\widehat{E_{di}}$, $i.e.$, $\widehat{E_{di}}$=$C_1 C_1 AC_{d} E_{i}^{128}$, where $d$ is set to be 3, 5, and 7. Subsequently, the generated features ($\widehat{E_{3i}}$, $\widehat{E_{5i}}$ and $\widehat{E_{7i}}$) are aggregated using a progressive manner that advantage is to make full use of diverse local features and strengthen their correlation. The whole process is as follows:
\begin{equation}
	\begin{split}
		& L_{i}^{1}=C_3\Phi(P_{3i}^{5},P_{3i}^{7},\widehat{E_{3i}}))+E_{i}^{128}, \\
		& P_{3i}^{k}=C_3\Phi(\widehat{E_{3i}},\widehat{E_{ki}},E_{i}^{128}), k=5,7,
	\end{split}   
\end{equation}
where $C_3$ denotes 3$\times$3 convolution. $\Phi(\cdot,\cdot,\cdot)$ and ``$+$'' represent concatenation and element-wise addition. The first stage excavates spatial local information through atrous convolution, while in the second stage, we exploit depth-wise convolution with different kernel sizes to capture different local details. Technically, the feature $L_{i}^{1}$ from the first stage is taken as input, and then we utilize multiple $n\times n$ depth-wise convolutions ($DC_n$) followed by two 1$\times$1 convolutions ($C_1$) to obtain local information ($\widehat{L_{ni}^{1}}$) with different perspectives, that is, $\widehat{L_{ni}^{1}}$=$C_1C_1DC_nL_{i}^{1}, n=3,5,7.$ Similar to the first stage, these local features are aggregated in a progressive manner to generate feature $L_{i}^{2}$, which is defined as: 
\begin{equation}
	\begin{split}
		& L_{i}^{2}=C_3\Phi(Y_{3i}^{5},Y_{3i}^{7},\widehat{L_{3i}^{1}}))+L_{i}^{1}, \\
		& Y_{3i}^{k}=C_3\Phi(\widehat{L_{3i}^{1}},\widehat{L_{ki}^{1}},L_{i}^{1}), k=5,7,
	\end{split}   
\end{equation}
where $\Phi(\cdot,\cdot,\cdot)$, $C_3$, and ``$+$'' are the same as in Eq. (3). Finally, we also use residual connections to alleviate the vanishing gradient problem and generate the local feature $L_i$, $i.e.$, $L_{i}$=$C_{3}\Phi(L_{i}^{2},E_{i}^{128})+E_{i}^{128}$. Through the two-stage operations, the feature $L_i$ contains abundant local details.
 \begin{figure}[t]
	\centering\includegraphics[width=0.48\textwidth,height=2.5cm]{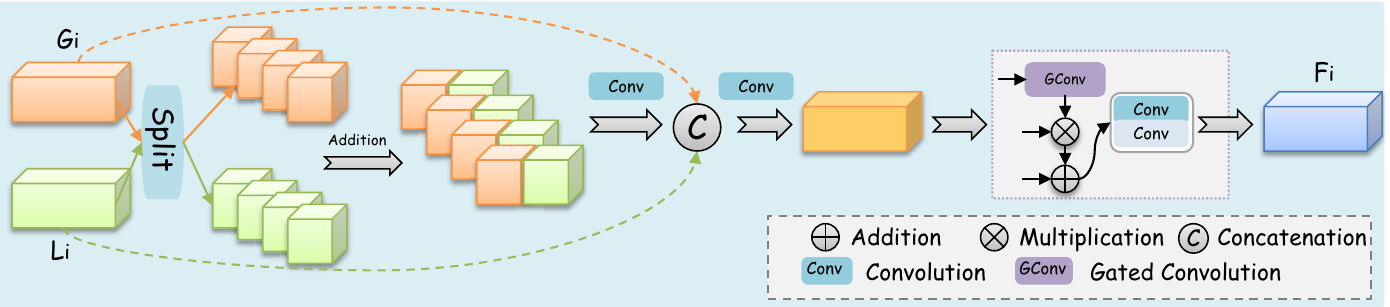}
	\captionsetup{font={small}, justification=raggedright}
	\caption{Details of the group-wise hybrid interaction module.}
	\label{Fig.4}
\end{figure}
\subsubsection{Group-wise hybrid interaction module}
Given a global feature $G_i$ and a local feature $L_i$, we propose a group-wise hybrid interaction module (GHIM) that aims to integrate global-local information through a grouping fusion with different channel spaces. Specifically, as shown in Fig. \ref{Fig.4}, we first split same-level $G_i$ and $L_i$ into four group features ($i.e.$, $\left\{ G_{i}^{m}\right \}_{m=1}^{4}$ and $\left\{ L_{i}^{m}\right \}_{m=1}^{4}$) of 32-channel and then perform aggregation optimization by utilizing element-wise addition and a 3$\times$3 convolution ($C_3$) for the four feature groups. Besides, we use a gated convolution ($GC_3$) for redundant information filtering to enhance the discrimination of features and perform a residual connection to generate feature $F_i$ with abundant global-local information. The proposed GHIM can be written as:
\begin{equation}
	\begin{split}
		& F_{i} = C_1C_3(GC_3\widetilde{F_{i}}\odot\widetilde{F_{i}}+\widetilde{F_{i}}), \\
		& \widetilde{F_{i}} = C_3\Phi (F_{i}^{1},F_{i}^{2},F_{i}^{3},F_{i}^{4},G_i,L_i), \\
		& \left \{ F_{i}^{m}\right\}_{m=1}^{4} = C_{3}(\left\{ L_{i}^{m} \right\}_{m=1}^{4} +\left\{ G_{i}^{m}\right \}_{m=1}^{4}), 
	\end{split}     
\end{equation}
where ``$\odot$'' denotes element-wise multiplication. $GC_3$ is a gated convolution \cite{gated} with kernel size of 3$\times$3.  $\Phi(\cdot,\cdot,\cdot,\cdot,\cdot,\cdot)$ is the concatenation. $C_3$ and ``$+$'' are the same as in Eq. (3).
\begin{table*}[t]
	\renewcommand{\arraystretch}{0.8}
	\setlength{\tabcolsep}{4pt}
	\centering
	\caption{ Quantitative results on three COD benchmark datasets. The best two results are shown in \textbf{\color{red} red} and \textbf{\color{green} green} . The symbols ``$\uparrow$/$\downarrow$'' denote that a higher/lower value is better. Note that ``Ours-R'', ``Ours-S'', and ``Ours-P'' present ResNet-50 \cite{ResNet}, Swin Transformer \cite{Swin}, and Pyramid Vision Transformer \cite{PVT} as the backbone, respectively.}
	\resizebox*{\textwidth}{78mm}{
		\begin{tabular}{ccccccccccccccccc}
			\toprule[1pt]\toprule[1pt]
			\multicolumn{1}{c|}{\multirow{2}{*}{\normalsize Method}} & \multicolumn{1}{c|}{\multirow{2}{*}{\normalsize Publication}} & \multicolumn{5}{c|}{CAMO-Test (250 images)}                       & \multicolumn{5}{c|}{COD10K-Test (2026 images)}                    & \multicolumn{5}{c}{NC4K (4121 images)}       \\
			\multicolumn{1}{c|}{}                        & \multicolumn{1}{c|}{}                             & $MAE$ $\downarrow$   & $AF_m$ $\uparrow$   & $WF_m$  $\uparrow$   & $S_m$ $\uparrow$    & \multicolumn{1}{c|}{$E_m$ $\uparrow$}    & $MAE$ $\downarrow$   & $AF_m$ $\uparrow$   & $WF_m$ $\uparrow$   & $S_m$ $\uparrow$    & \multicolumn{1}{c|}{$E_m$ $\uparrow$}     & $MAE$ $\downarrow$   & $AF_m$ $\uparrow$   & $WF_m$ $\uparrow$   & $S_m$ $\uparrow$    & $E_m$ $\uparrow$    \\ \toprule[0.5pt]
			\multicolumn{17}{c}{Convolution backbone-based Method}                                                                                                                                                                                                             \\ \toprule[0.5pt]
			\rowcolor{blue!10}\multicolumn{1}{c|}{SINet\cite{COD10K}}                   & \multicolumn{1}{c|}{2020, CVPR}                    & 0.100 & 0.709 & 0.606 & 0.751 & \multicolumn{1}{c|}{0.835} & 0.051 & 0.593 & 0.551 & 0.770 & \multicolumn{1}{c|}{0.797} & 0.058 & 0.768 & 0.723 & 0.807 & 0.883 \\ 
			\multicolumn{1}{c|}{PFNe\cite{PFNet}}                   & \multicolumn{1}{c|}{2021, CVPR}                    & 0.085 & 0.751 & 0.695 & 0.782 & \multicolumn{1}{r|}{0.855} & 0.040 & 0.676 & 0.660 & 0.798 & \multicolumn{1}{c|}{0.868} & 0.053 & 0.779 & 0.745 & 0.828 & 0.894 \\ 
			\rowcolor{blue!10}\multicolumn{1}{c|}{MGL-S\cite{MGL}}                   & \multicolumn{1}{c|}{2021, CVPR}                    & 0.089 & 0.733 & 0.664 & 0.772 & \multicolumn{1}{c|}{0.850} & 0.037 & 0.667 & 0.655 & 0.808 & \multicolumn{1}{c|}{0.851} & 0.055 & 0.771 & 0.731 & 0.828 & 0.885 \\ 
			\multicolumn{1}{c|}{LSR\cite{NC4K}}                     & \multicolumn{1}{c|}{2021, CVPR}                    & 0.080 & 0.756 & 0.696 & 0.787 & \multicolumn{1}{c|}{0.859} & 0.037 & 0.699 & 0.673 & 0.802 & \multicolumn{1}{c|}{0.883} & 0.048 & 0.802 & 0.766 & 0.839 & 0.904 \\ 
			\rowcolor{blue!10}\multicolumn{1}{c|}{JSOCOD\cite{JSOCOD}}                     & \multicolumn{1}{c|}{2021, CVPR}                    & 0.073 & 0.779 & 0.728 & 0.800 & \multicolumn{1}{c|}{0.872} & 0.035 & 0.705 & 0.684 & 0.807 & \multicolumn{1}{c|}{0.882} & 0.047 & 0.803 & 0.771 & 0.841 & 0.906 \\ 
			\multicolumn{1}{c|}{UGTR\cite{UGTR}}                    & \multicolumn{1}{c|}{2021, ICCV}                    & 0.086 & 0.748 & 0.684 & 0.784 & \multicolumn{1}{c|}{0.858} & 0.036 & 0.671 & 0.666 & 0.815 & \multicolumn{1}{c|}{0.850} & 0.052 & 0.778 & 0.747 & 0.839 & 0.888 \\ 
			\rowcolor{blue!10}\multicolumn{1}{c|}{C2FNet\cite{C2FNet}}                  & \multicolumn{1}{c|}{2021, IJCAI}                   & 0.080 & 0.764 & 0.719 & 0.796 & \multicolumn{1}{c|}{0.865} & 0.036 & 0.703 & 0.686 & 0.811 & \multicolumn{1}{c|}{0.886} & 0.049 & 0.788 & 0.762 & 0.838 & 0.901 \\ 
			\multicolumn{1}{c|}{TINet\cite{TINet}}                   & \multicolumn{1}{c|}{2021, AAAI}                    & 0.087 & 0.730 & 0.678 & 0.781 & \multicolumn{1}{c|}{0.847} & 0.043 & 0.652 & 0.635 & 0.791 & \multicolumn{1}{c|}{0.848} & 0.055 & 0.766 & 0.734 & 0.828 & 0.882 \\ 
			\rowcolor{blue!10}\multicolumn{1}{c|}{ERRNet\cite{ERRNet}}                  & \multicolumn{1}{c|}{2022, PR}                      & 0.085 & 0.731 & 0.679 & 0.778 & \multicolumn{1}{c|}{0.855} & 0.043 & 0.646 & 0.630 & 0.785 & \multicolumn{1}{c|}{0.845} & 0.054 & 0.769 & 0.737 & 0.826 & 0.892 \\ 
			\multicolumn{1}{c|}{CubeNet\cite{CubeNet}}                 & \multicolumn{1}{c|}{2022, PR}                      & 0.085 & 0.734 & 0.682 & 0.788 & \multicolumn{1}{c|}{0.852} & 0.041 & 0.671 & 0.644 & 0.793 & \multicolumn{1}{c|}{0.863} & -     & -     & -     & -     & -     \\ 
			\rowcolor{blue!10}\multicolumn{1}{c|}{PreyNet\cite{PreyNet}}                 & \multicolumn{1}{c|}{2022, MM}                      & 0.077 & 0.764 & 0.708 & 0.789 & \multicolumn{1}{c|}{0.856} & \textbf{\color{green}0.034} & \textbf{\color{red}0.731} & 0.697 & 0.810 & \multicolumn{1}{c|}{\textbf{\color{red}0.894}} & -     & -     & -     & -     & -     \\ 
			\multicolumn{1}{c|}{FAPNet\cite{FAPNet}}                  & \multicolumn{1}{c|}{2022, TIP}                     & 0.076 & \textbf{\color{green}0.776} & \textbf{\color{green}0.734} & \textbf{\color{green}0.815} & \multicolumn{1}{c|}{\textbf{\color{green}0.877}} & 0.036 & 0.707 & 0.694 & \textbf{\color{green}0.820} & \multicolumn{1}{c|}{0.875} & \textbf{\color{green}0.047} & 0.804 & \textbf{\color{green}0.775} & \textbf{\color{green}0.850} & 0.903 \\ 
			\rowcolor{blue!10}\multicolumn{1}{c|}{BSANet\cite{BSANet}}                  & \multicolumn{1}{c|}{2022, AAAI}                    & 0.079 & 0.768 & 0.717 & 0.794 & \multicolumn{1}{c|}{0.866} & \textbf{\color{green}0.034} & 0.724 & \textbf{\color{green}0.699} & 0.815 & \multicolumn{1}{c|}{\textbf{\color{red}0.894}} & 0.048 & \textbf{\color{green}0.805} & 0.771 & 0.841 & \textbf{\color{green}0.906} \\ 
			\multicolumn{1}{c|}{SegMaR\cite{SegMaR}}                & \multicolumn{1}{c|}{2022, CVPR}                    & \textbf{\color{green}0.072} & 0.772 & 0.728 & 0.808 & \multicolumn{1}{c|}{0.870} & 0.035 & 0.699 & 0.682 & 0.811 & \multicolumn{1}{c|}{0.881} & -     & -     & -     & -     & -     \\ 
			\rowcolor{blue!10}\multicolumn{1}{c|}{PUENet\cite{PUENet}}                & \multicolumn{1}{c|}{2023, TIP}                    & 0.080 & 0.762 & - & 0.794 & \multicolumn{1}{c|}{0.857} & 0.035 & 0.727 & - & 0.813 & \multicolumn{1}{c|}{0.887} & 0.050     & 0.798     & -     & 0.836     & 0.892     \\
   
            \multicolumn{1}{c|}{MRRNet}\cite{TNNLS3}                & \multicolumn{1}{c|}{2023, TNNLS}                    & 0.076 & 0.766 & 0.728 & 0.811 & \multicolumn{1}{c|}{0.870} & 0.036 & 0.695 & 0.692 & \textbf{\color{green}0.820} & \multicolumn{1}{c|}{0.869} & 0.049     & 0.788     & 0.766     & 0.847     & 0.894     \\ \hline
   
			\rowcolor{blue!10}\multicolumn{1}{c|}{Ours-R}                 & \multicolumn{1}{c|}-                                                 & \textbf{\color{red}0.069} & \textbf{\color{red}0.788} & \textbf{\color{red}0.748} & \textbf{\color{red}0.816} & \multicolumn{1}{c|}{\textbf{\color{red}0.882}} & \textbf{\color{red}0.033} & \textbf{\color{green}0.729} & \textbf{\color{red}0.714} & \textbf{\color{red}0.822} & \multicolumn{1}{c|}{\textbf{\color{green}0.891}} & \textbf{\color{red}0.043} & \textbf{\color{red}0.816} & \textbf{\color{red}0.791} & \textbf{\color{red}0.852} & \textbf{\color{red}0.914} \\ \toprule[0.5pt]
			\multicolumn{17}{c}{Transformer backbone-based Method}                                                                                                                                                                                                            \\ \toprule[0.5pt]
			\rowcolor{blue!8}\multicolumn{1}{c|}{VST\cite{VST}}                     & \multicolumn{1}{c|}{2021, ICCV}                    & 0.081 & 0.753 & 0.713 & 0.808 & \multicolumn{1}{c|}{0.853} & 0.037 & 0.721 & 0.698 & 0.817 & \multicolumn{1}{c|}{0.882} & 0.048 & 0.801 & 0.768 & 0.844 & 0.899 \\ 
			\multicolumn{1}{c|}{EVP\cite{EVP}}                     & \multicolumn{1}{c|}{2023, CVPR}                    & 0.067 & 0.800 & 0.762 & 0.831 & \multicolumn{1}{c|}{0.896} & 0.032 & 0.708 & 0.726 & 0.835 & \multicolumn{1}{c|}{0.877} & -     & -     & -     & -     & -     \\ 
			\rowcolor{blue!8}\multicolumn{1}{c|}{FPNet\cite{FPNet}}                     & \multicolumn{1}{c|}{2023, MM}                    & 0.056 & {0.838} & {0.802} & 0.851 & \multicolumn{1}{c|}{0.912} & 0.029 & {0.765} & {0.755} & {0.847} & \multicolumn{1}{c|}{{0.909}} & -     & -     & -     & -     & -     \\ 
			\multicolumn{1}{c|}{FSPNet\cite{FSPNet}}                  & \multicolumn{1}{c|}{2023, CVPR}                    & {0.050} & 0.829 & 0.799 & {0.855} & \multicolumn{1}{c|}{{0.919}} & {0.026} & 0.736 & 0.735 & {0.847} & \multicolumn{1}{c|}{0.900} & {0.035} & {0.826} & {0.816} & {0.878} & {0.923} \\ \hline
   
            \rowcolor{blue!8}\multicolumn{1}{c|}{Ours-P}                  & \multicolumn{1}{c|}{-}                    & \textbf{\color{green}0.042} & \textbf{\color{green}0.861} & \textbf{\color{green}0.847} & \textbf{\color{green}0.878} & \multicolumn{1}{c|}{\textbf{\color{green}0.939}} & \textbf{\color{red}0.022} & \textbf{\color{red}0.796} & \textbf{\color{red}0.794} & \textbf{\color{red}0.866} & \multicolumn{1}{c|}{\textbf{\color{red}0.930}} & \textbf{\color{green}0.031} & \textbf{\color{green}0.856} & \textbf{\color{green}0.846} & \textbf{\color{green}0.884} & \textbf{\color{green}0.939} \\ 
			\rowcolor{blue!8}\multicolumn{1}{c|}{Ours-S}                 & \multicolumn{1}{c|}-                                                 & \textbf{\color{red}0.038} & \textbf{\color{red}0.864} & \textbf{\color{red}0.849} & \textbf{\color{red}0.880} & \multicolumn{1}{c|}{\textbf{\color{red}0.940}} & \textbf{\color{green}0.023} & \textbf{\color{green}0.784} & \textbf{\color{green}0.783} & \textbf{\color{green}0.860} & \multicolumn{1}{c|}{\textbf{\color{green}0.929}} & \textbf{\color{red}0.030} & \textbf{\color{red}0.858} & \textbf{\color{red}0.847} & \textbf{\color{red}0.886} & \textbf{\color{red}0.942} \\ \toprule[1pt]\toprule[1pt]
		\end{tabular}}
        \label{TABLE2}
	\end{table*}
\subsection{Adjacent Reverse Decoder}
After obtaining the optimized feature $F_i$, we need to decode the feature $F_i$ to generate the predicted map $D_i$. Existing COD methods \cite{BSANet,TINet} usually utilize the feature pyramid network (FPN) \cite{FPN} to aggregate features at different levels. However, there are two limitations to the FPN decoder, \textbf{1)} the semantic information is gradually diluted as the hierarchy becomes deeper; \textbf{2)} a single concatenation does not adequately aggregate complementary information ($i.e.$, semantics from high-level features and details from low-level features), which is not conducive to the segmentation of camouflaged objects. Therefore, we propose an adjacent reverse decoder (ARD) that integrates adverse information through cross-layer aggregation and reverse optimization to alleviate the above problem. 
\begin{figure}[]
		\centering\includegraphics[width=0.48\textwidth,height=2.8cm]{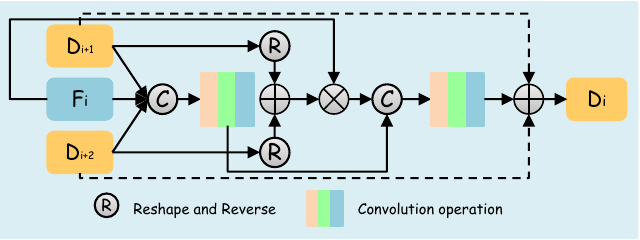}
		\captionsetup{font={small}, justification=raggedright}
		\caption{Details of the adjacent reverse decoder. }
		\label{Fig.5}
\end{figure}

Specifically, we first feed features $E_5$ and $G_5$ into the MTB, which incorporates a series of dimension reduction convolutions to generate a coarse predicted map with 1-channel $D_6$, $i.e.$, $D_6$ = $C_3C_1MTB(\Phi(E_5, G_5))$, where $\Phi(\cdot,\cdot)$ denotes the concatenation. Subsequently, we input feature maps from different layer $D_{i+1}$, $D_{i+2}$ and $F_i$ into the ARD to generate a feature map $D_i$, as shown in Fig. \ref{Fig.5}. Technically, $D_{i+1}$ and $D_{i+2}$ are performed upsampling and dimension expansion to ensure that its channel and size are the same as $F_i$ in the concatenation progress to generate feature $F_{i}^{c}$, $i.e.$, $F_{i}^{c}= \varrho(\Phi (F_i,\mho(D_{i+1}),\mho(D_{i+2})))$, where $\varrho(\cdot)$ denotes multiple convolution operations, $\mho(\cdot)$ represents pixelshuffle upsampling and dimension extension operations. Furthermore, we generate a reversed attention map ($\Lambda$) by using a reversed attention acting on features $D_{i+1}$ and $D_{i+2}$ for optimizing feature $F_i$ to generate optimized feature $F_{i}^{r}$. Finally, the feature $F_{i}^{c}$ and $F_{i}^{r}$ are concatenated and dimensionally reduced, and then two feature maps $D_{i+1}$ and $D_{i+2}$ are added to generate the final feature map $D_i$, that is, 
\begin{equation}
	\begin{split}
			& D_i = C_3(\Phi(F_{i}^{c},F_{i}^{r}))+D_{i+1}+D_{i+2}, \\
			& F_{i}^{r} = \Lambda F_i,  \Lambda = \mho(RA(D_{i+1}))+\mho(RA(D_{i+2})), \\
	\end{split}  
\end{equation}
where $C_3$ denotes the convolution with a kernal size of 3$\times$3, $RA(\cdot)$ is a reverse attention mechanism. Note that $i+1\le6$ and $i+2\le6$, since the feature $D_5$ is required only $D_6$ and $F_5$ as inputs. The purpose of ARD is to excavate and exploit the potential significant information of features with different layers that can be used to distinguish subtle differences between camouflaged objects and surroundings.
\subsection{Loss Functions}
Similar to these methods \cite{BSANet,FAPNet,FSPNet}, we combine the weighted binary cross-entry loss ($\Im_{BCE}^{w}$) and the weighted intersection-over-union loss ($\Im_{IoU}^{w}$) though ground truths for supervised training. The loss function $\Im_{all}$ is formulated as:
\begin{equation}
	\Im_{all} = \sum_{i=2}^{6} (\Im_{BCE}^{w}(S|D_i|,G)+\Im_{IoU}^{w}(S|D_i|,G)),  
\end{equation}
where $S|\cdot|$ denotes the sigmoid function, and $G$ presents the ground truth.

The weighted binary cross-entry loss function ($\Im_{BCE}^{w}$) has been widely used in the field of image segmentation tasks, which is defined as follows:
\begin{small} 
	\begin{equation}
		\begin{aligned}
			\Im_{BCE}^{w}=-\frac{\!\sum_{(i,j)}^{HW}\!(1\!+\!w_{ij})(G_{ij}logP_{ij}\!+\!(1\!-\!G_{ij})(1\!-\!logP_{ij}))}{\!\sum_{(i,j)}^{HW}(1\!+\!w_{ij})\!},
		\end{aligned}
	\end{equation}
\end{small}where $w_{ij}$ presents the weight value of the pixel $(i, j)$. $P_{ij}$ and $G_{ij}$ denote the method prediction and ground truth of the pixel $(i, j)$ to be camouflaged objects. To optimize the global structure, we introduce the weighted intersection-over-union (IoU) loss \cite{IoU}, which computes the complete structural similarity. $\Im_{IoU}^{w}$ is given as follows:
\begin{small}
\begin{equation}
	\begin{aligned}
		\Im_{IoU}^{w}=1-\frac{{\displaystyle\sum_{(i,j)}^{HW}}\;(1+w_{ij})G_{ij} P_{ij}}{{\displaystyle\sum_{(i,j)}^{HW}}\;(1+w_{ij})(G_{ij}+P_{ij}-G_{ij} P_{ij})}.
	\end{aligned}
\end{equation}
\end{small}
\begin{figure*}[]
	\scriptsize
	\centering
	\begin{tabular}{ccc}		
		\includegraphics[width=0.28\textwidth,height=3.5cm]{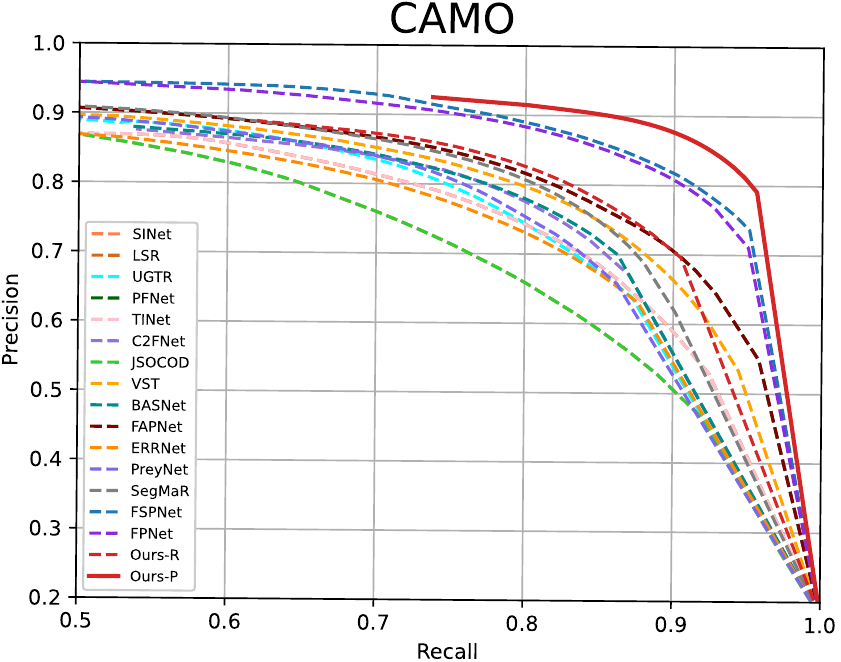}\
		&\includegraphics[width=0.28\textwidth,height=3.5cm]{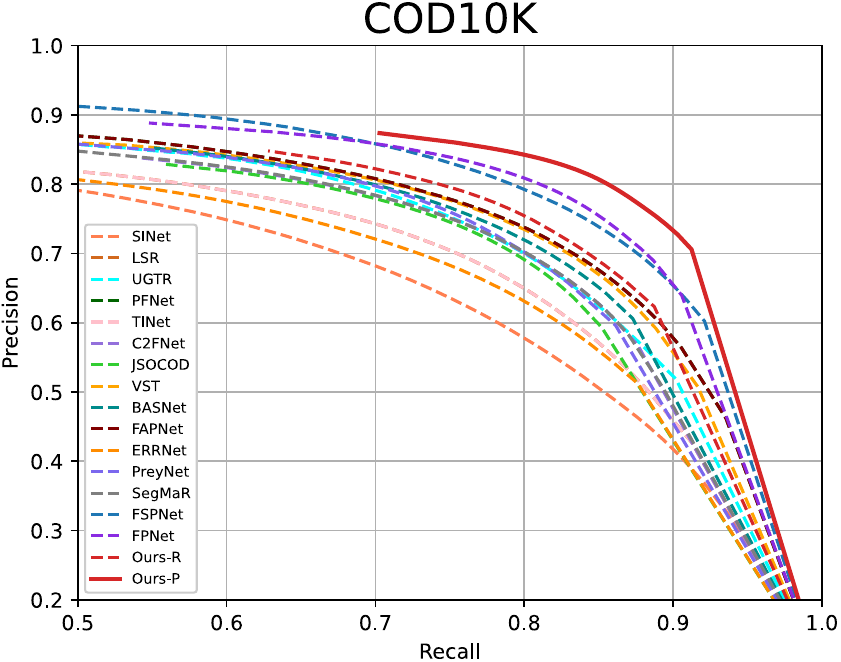}\
		&\includegraphics[width=0.28\textwidth,height=3.5cm]{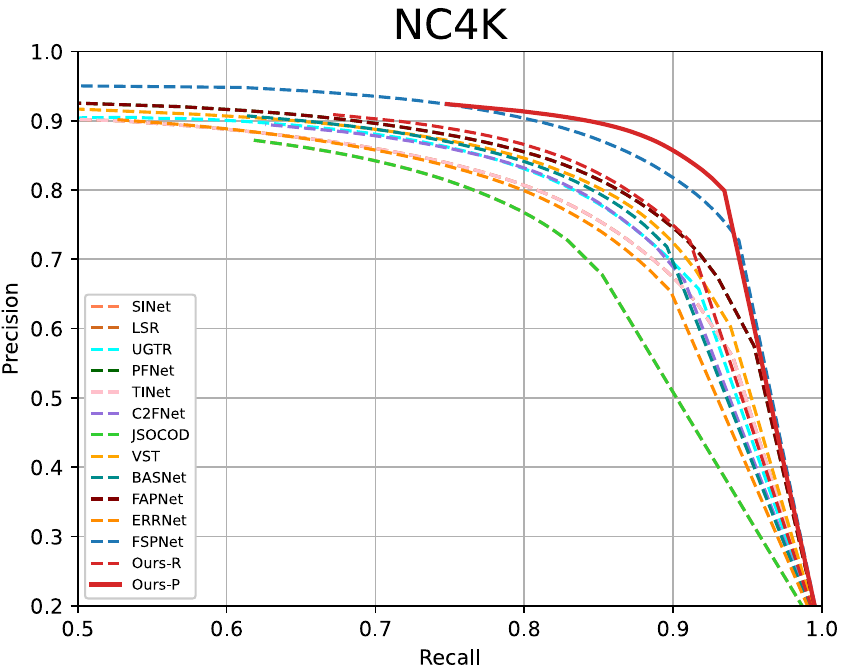}\\	
	\end{tabular}
	\begin{tabular}{ccc}		
		\includegraphics[width=0.28\textwidth,height=3.5cm]{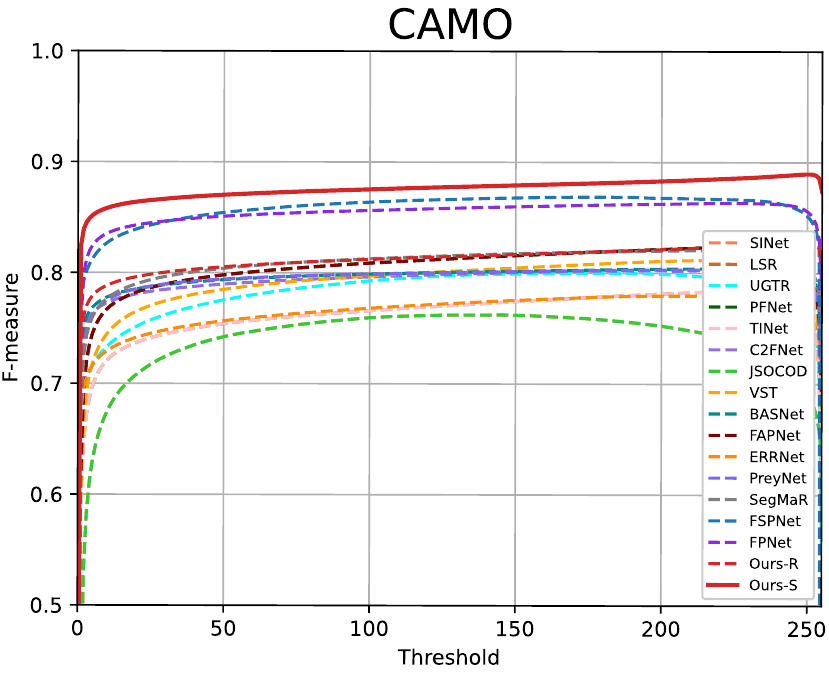}\
		&\includegraphics[width=0.28\textwidth,height=3.5cm]{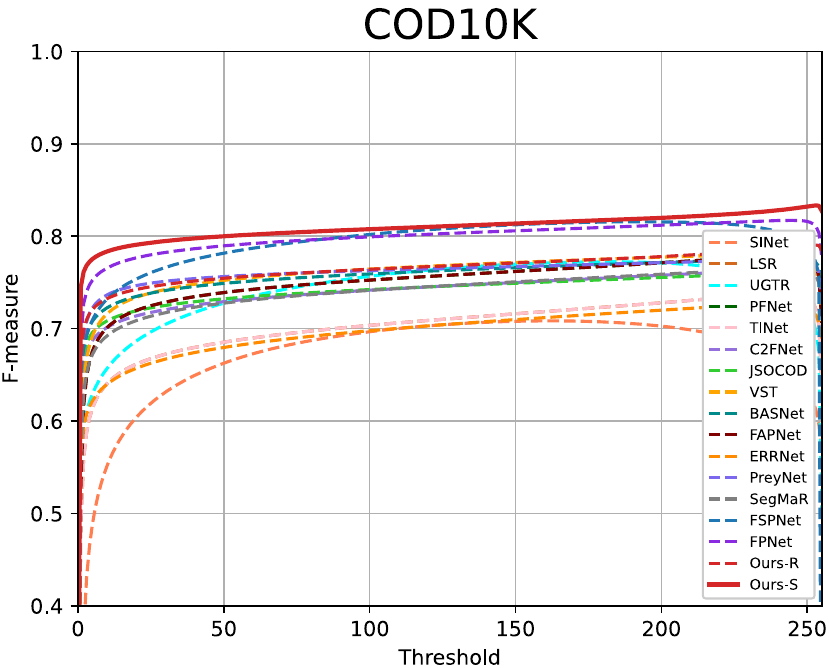}\
		&\includegraphics[width=0.28\textwidth,height=3.5cm]{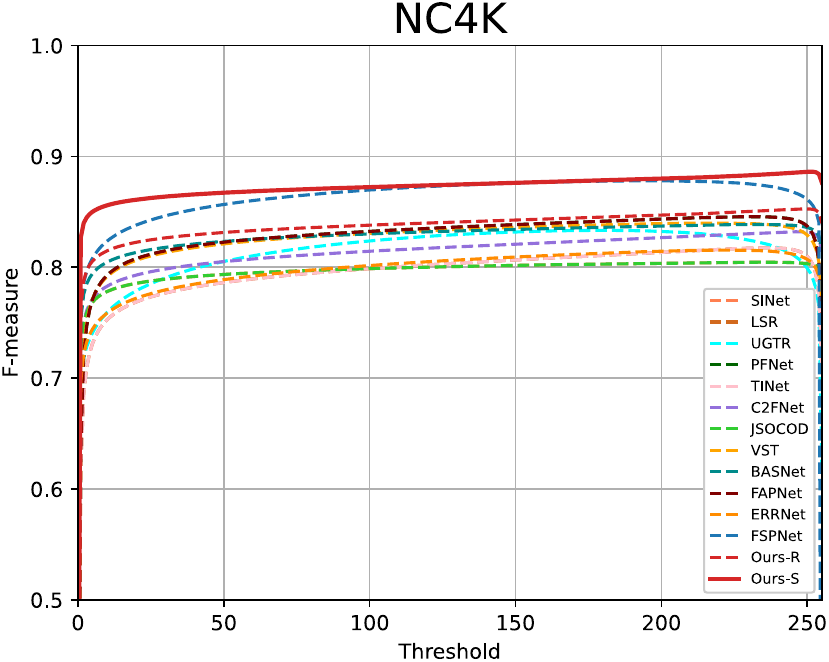}\\
	\end{tabular}		
	\captionsetup{font={small}, justification=raggedright}
	\caption{Quantitative resluts of the $PR$ and $F_m$ curve for GLCONet and other COD methods on three COD datasets.}
	\label{Fig.6}
\end{figure*}
\begin{figure*}[t]
	\centering\includegraphics[width=0.90\textwidth,height=6.8cm]{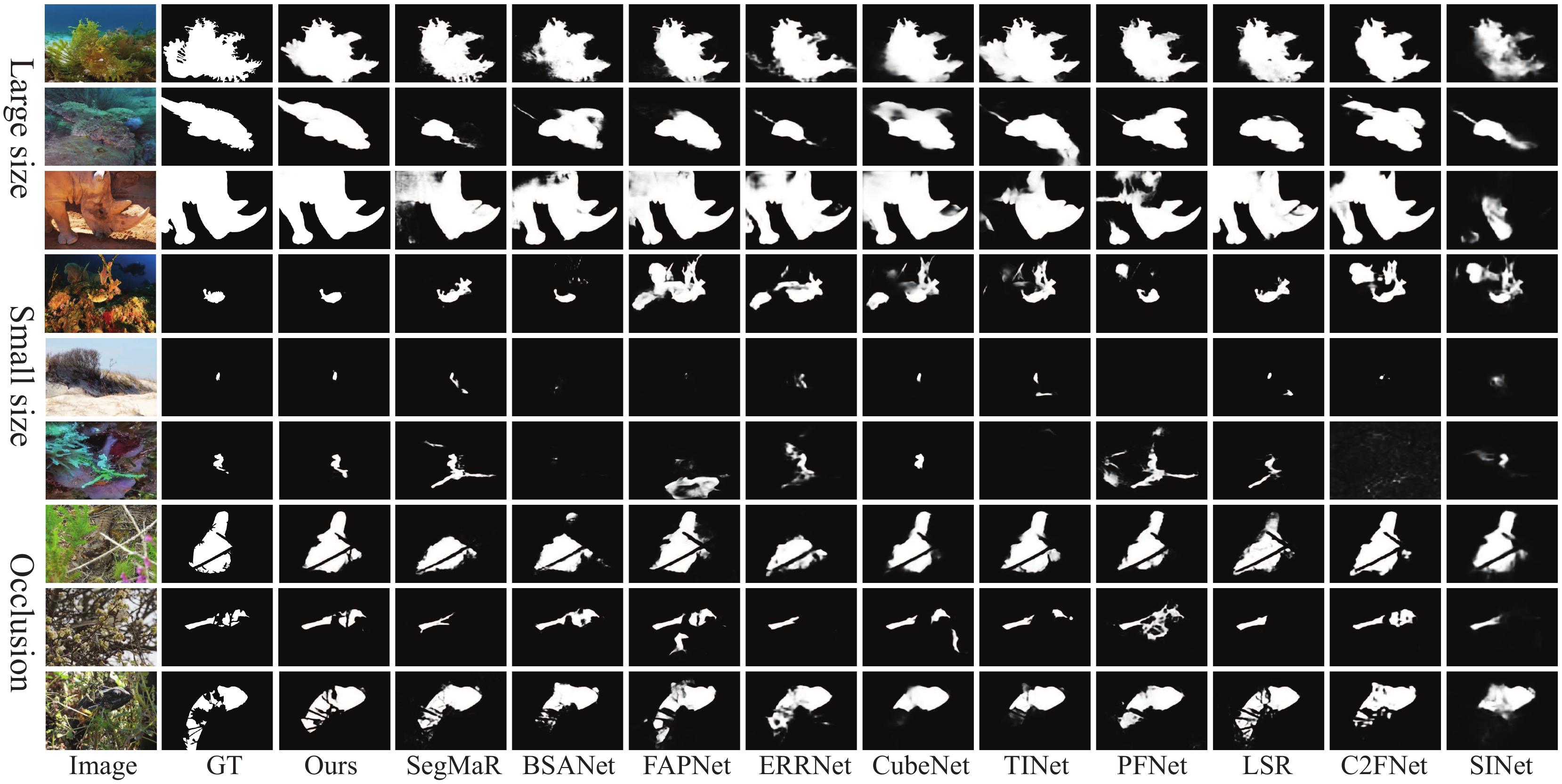}
	\captionsetup{font={small}, justification=raggedright}
	\caption{Qualitative comparisons of the proposed GLCONet method and 10 COD methods.}
	\label{Fig.7}		
\end{figure*}
\section{Experiment}
With our method's powerful capture of local details and long-range dependencies, and ARD's efficient decoding of diverse information from different layers, our method shows strong competitiveness across multiple datasets. In this section, we first introduce some experiment settings and then show comparison results and ablation studies of our GLCONet.
\begin{table*}[t]
	\renewcommand{\arraystretch}{0.8}
	\setlength{\tabcolsep}{3pt}
	\centering
	\caption{ Quantitative results on the COD10K dataset of four categories ($i.e.$, ``large objects'', ``medium objects'', ``small objects'', and ``occluded objects''). The best two results are shown in \textbf{\color{red} red} and \textbf{\color{green} green} . The symbols ``$\uparrow$/$\downarrow$'' denote that a higher/lower value is better.}
	\resizebox*{\textwidth}{70mm}{
		\begin{tabular}{ccccccccccccccccccccc}
			\toprule[1pt]\toprule[1pt]
			\multicolumn{1}{c|}{\multirow{2}{*}{\normalsize Methods}} & \multicolumn{5}{c|}{Large objects (165 images)}                    & \multicolumn{5}{c|}{Medium objects (543 images)}                   & \multicolumn{5}{c|}{Small objects (853 images)}                    & \multicolumn{5}{c}{Occluded objects(465 images)} \\
			\multicolumn{1}{c|}{}                        & $MAE$ $\downarrow$   & $AF_m$ $\uparrow$   & $WF_m$  $\uparrow$   & $S_m$ $\uparrow$    & \multicolumn{1}{c|}{$E_m$ $\uparrow$}    & $MAE$ $\downarrow$   & $AF_m$ $\uparrow$   & $WF_m$ $\uparrow$   & $S_m$ $\uparrow$    & \multicolumn{1}{c|}{$E_m$ $\uparrow$}     & $MAE$ $\downarrow$   & $AF_m$ $\uparrow$   & $WF_m$ $\uparrow$   & $S_m$ $\uparrow$    & \multicolumn{1}{c|}{$E_m$ $\uparrow$}    & $MAE$ $\downarrow$   & $AF_m$ $\uparrow$   & $WF_m$  $\uparrow$   & $S_m$ $\uparrow$    & $E_m$ $\uparrow$    \\ \toprule[0.5pt]
			\multicolumn{21}{c}{Convolution backbone-based COD Method}                                                                    \\ \toprule[0.5pt]
			\multicolumn{1}{c|}{SINet$_{20}$\cite{COD10K}}                   & 0.137 & 0.837 & 0.719 & 0.776 & \multicolumn{1}{c|}{0.827} & 0.055 & 0.773 & 0.689 & 0.827 & \multicolumn{1}{c|}{0.925} & 0.032 & 0.467 & 0.468 & 0.785 & \multicolumn{1}{c|}{0.722} & 0.052  & 0.528  & 0.482  & 0.729 & 0.774 \\ 
			\rowcolor{blue!8}\multicolumn{1}{c|}{PFNet$_{21}$\cite{PFNet}}                   & 0.098 & 0.872 & 0.812 & 0.829 & \multicolumn{1}{c|}{0.873} & 0.040 & 0.816 & 0.782 & 0.858 & \multicolumn{1}{c|}{0.938} & 0.026 & 0.582 & 0.591 & 0.778 & \multicolumn{1}{c|}{0.830} & 0.043  & 0.615  & 0.590  & 0.753 & 0.854 \\ 
			\multicolumn{1}{c|}{MGL-S$_{21}$\cite{MGL}}                   & 0.098 & 0.876 & 0.807 & 0.833 & \multicolumn{1}{c|}{0.865} & 0.038 & 0.828 & 0.785 & 0.869 & \multicolumn{1}{c|}{0.942} & 0.023 & 0.556 & 0.581 & 0.788 & \multicolumn{1}{c|}{0.797} & 0.039  & 0.609  & 0.586  & 0.766 & 0.839 \\ 
			\rowcolor{blue!8}\multicolumn{1}{c|}{LSR$_{21}$\cite{NC4K}}                     & 0.094 & 0.879 & 0.821 & 0.837 & \multicolumn{1}{c|}{0.869} & 0.038 & 0.828 & 0.793 & 0.863 & \multicolumn{1}{c|}{0.939} & 0.024 & 0.614 & 0.609 & 0.784 & \multicolumn{1}{c|}{0.857} & 0.041  & 0.639  & 0.598  & 0.754 & 0.869 \\ 
			\multicolumn{1}{c|}{UGTR$_{21}$\cite{UGTR}}                    & 0.087 & 0.889 & 0.831 & 0.854 & \multicolumn{1}{c|}{0.879} & 0.036 & 0.832 & 0.796 & 0.877 & \multicolumn{1}{c|}{0.945} & 0.024 & 0.556 & 0.585 & 0.789 & \multicolumn{1}{c|}{0.789} & \textbf{\color{green}0.038}  & 0.617  & 0.605  & \textbf{\color{green}0.776} & 0.840 \\ 
			\rowcolor{blue!8}\multicolumn{1}{l|}{JSOCOD$_{21}$\cite{JSOCOD}}                  & 0.084 & 0.892 & 0.838 & 0.846 & \multicolumn{1}{c|}{0.885} & \textbf{\color{green}0.034} & 0.842 & \textbf{\color{green}0.814} & 0.872 & \multicolumn{1}{c|}{0.946} & 0.024 & 0.618 & 0.616 & 0.786 & \multicolumn{1}{c|}{0.850} & 0.040  & 0.638  & 0.602  & 0.755 & 0.867 \\ 
			\multicolumn{1}{c|}{C2FNet$_{21}$\cite{C2FNet}}                  & 0.091 & 0.880 & 0.826 & 0.840 & \multicolumn{1}{c|}{0.880} & 0.037 & 0.826 & 0.800 & 0.865 & \multicolumn{1}{c|}{0.944} & 0.024 & 0.623 & 0.623 & 0.792 & \multicolumn{1}{c|}{0.857} & \textbf{\color{green}0.038}  & 0.642  & 0.619  & 0.771 & 0.872 \\ 
			\rowcolor{blue!8}\multicolumn{1}{c|}{TINet$_{21}$\cite{TINet}}                   & 0.099 & 0.867 & 0.805 & 0.830 & \multicolumn{1}{c|}{0.872} & 0.045 & 0.745 & 0.797 & 0.763 & \multicolumn{1}{c|}{0.852} & 0.030 & 0.548 & 0.555 & 0.767 & \multicolumn{1}{c|}{0.796} & 0.045  & 0.600  & 0.573  & 0.751 & 0.842 \\ 
			\multicolumn{1}{c|}{ERRNet$_{22}$\cite{ERRNet}}                  & 0.103 & 0.864 & 0.798 & 0.822 & \multicolumn{1}{c|}{0.865} & 0.045 & 0.791 & 0.757 & 0.846 & \multicolumn{1}{c|}{0.930} & 0.029 & 0.547 & 0.557 & 0.763 & \multicolumn{1}{c|}{0.795} & 0.047  & 0.582  & 0.557  & 0.739 & 0.830 \\ 
			\rowcolor{blue!8}\multicolumn{1}{c|}{CubeNet$_{22}$\cite{CubeNet}}                 & 0.098 & 0.871 & 0.802 & 0.830 & \multicolumn{1}{c|}{0.863} & 0.043 & 0.815 & 0.770 & 0.855 & \multicolumn{1}{c|}{0.937} & 0.028 & 0.574 & 0.573 & 0.771 & \multicolumn{1}{c|}{0.824} & 0.044  & 0.610  & 0.571  & 0.745 & 0.846 \\ 
			\multicolumn{1}{c|}{PreyNet$_{22}$\cite{PreyNet}}                 & 0.092 & 0.886 & 0.826 & 0.840 & \multicolumn{1}{c|}{0.875} & \textbf{\color{green}0.034} & \textbf{\color{green}0.852} & \textbf{\color{green}0.814} & 0.871 & \multicolumn{1}{c|}{0.940} & \textbf{\color{red}0.021} & \textbf{\color{red}0.659} & \textbf{\color{green}0.641} & 0.796 & \multicolumn{1}{c|}{\textbf{\color{red}0.879}} & \textbf{\color{green}0.038}  & \textbf{\color{green}0.665}  & 0.615  & 0.756 & 0.872 \\ 
			\rowcolor{blue!8}\multicolumn{1}{c|}{FAPNet$_{22}$\cite{FAPNet}}                  & \textbf{\color{green}0.077} & \textbf{\color{green}0.897} & \textbf{\color{green}0.854} & \textbf{\color{red}0.864} & \multicolumn{1}{c|}{\textbf{\color{green}0.890}} & 0.036 & 0.839 & 0.812 & {0.877} & \multicolumn{1}{c|}{0.944} & 0.025 & 0.625 & 0.631 & \textbf{\color{green}0.798} & \multicolumn{1}{c|}{0.838} & 0.040  & 0.637  & 0.616  & {0.775} & 0.858 \\ 
			\multicolumn{1}{c|}{BSANet$_{22}$\cite{BSANet}}                  & 0.086 & 0.890 & 0.836 & 0.845 & \multicolumn{1}{c|}{0.886} & 0.035 & 0.844 & 0.812 & 0.871 & \multicolumn{1}{c|}{\textbf{\color{green}0.948}} & \textbf{\color{green}0.022} & \textbf{\color{green}0.649} & 0.640 & 0.797 & \multicolumn{1}{c|}{\textbf{\color{green}0.870}} & \textbf{\color{red}0.037}  & 0.659  & {0.629}  & 0.772 & \textbf{\color{red}0.878} \\ 
			\rowcolor{blue!8}\multicolumn{1}{c|}{SegMaR$_{22}$\cite{SegMaR}}                & 0.080 & 0.888 & 0.843 & 0.857 & \multicolumn{1}{c|}{0.884} & 0.036 & 0.835 & 0.807 & 0.874 & \multicolumn{1}{c|}{0.946} & 0.024 & 0.609 & 0.611 & 0.788 & \multicolumn{1}{c|}{0.847} & 0.040  & 0.639  & 0.608  & 0.764 & 0.867 \\ 
            \multicolumn{1}{c|}{MRRNet$_{23}$}\cite{TNNLS3}                & 0.083 & 0.887 & 0.841 & 0.856 & \multicolumn{1}{c|}{\textbf{\color{green}0.890}} & 0.035 & 0.836 & 0.811 & \textbf{\color{green}0.879} & \multicolumn{1}{c|}{0.947} & 0.025 & 0.598 & 0.621 & 0.796 & \multicolumn{1}{c|}{0.820} & 0.040  & 0.641  & \textbf{\color{green}0.630}  & \textbf{\color{red}0.781} & 0.859 \\ \hline
			\rowcolor{blue!8}\multicolumn{1}{c|}{Ours-R}                 & \textbf{\color{red}0.076} & \textbf{\color{red}0.904} & \textbf{\color{red}0.860} & \textbf{\color{green}0.860} & \multicolumn{1}{c|}{\textbf{\color{red}0.901}} & \textbf{\color{red}0.031} & \textbf{\color{red}0.857} & \textbf{\color{red}0.836} & \textbf{\color{red}0.884} & \multicolumn{1}{c|}{\textbf{\color{red}0.954}} & 0.023 & 0.647 & \textbf{\color{red}0.646} & \textbf{\color{red}0.800} & \multicolumn{1}{c|}{0.856} & \textbf{\color{green}0.038}  & \textbf{\color{red}0.669}  & \textbf{\color{red}0.643}  & \textbf{\color{green}0.776} & \textbf{\color{green}0.877} \\ \toprule[0.5pt]
			\multicolumn{21}{c}{Transformer backbone-based COD Method}                                                                                                                                                                                                                    \\ \toprule[0.5pt]
			\rowcolor{blue!8}\multicolumn{1}{c|}{VST$_{21}$\cite{VST}}                     & 0.079 & 0.894 & 0.850 & 0.865 & \multicolumn{1}{c|}{0.889} & 0.037 & 0.846 & 0.811 & 0.877 & \multicolumn{1}{c|}{0.948} & 0.026 & 0.644 & 0.639 & 0.798 & \multicolumn{1}{c|}{0.851} & 0.043  & 0.654  & 0.617  & 0.765 & 0.859 \\ 
			\multicolumn{1}{c|}{FPNet$_{23}$\cite{FPNet}}                     & 0.072 & {0.911} & 0.873 & 0.876 & \multicolumn{1}{c|}{{0.902}} & {0.027} & 0.870 & {0.862} & {0.902} & \multicolumn{1}{c|}{0.959} & 0.021 & {0.682} & {0.686} & 0.822 & \multicolumn{1}{c|}{{0.880}} & {0.029}  & {0.732}  & {0.714}  & \textbf{\color{green}0.818} & {0.906} \\ 
			\rowcolor{blue!8}\multicolumn{1}{c|}{EVP$_{23}$\cite{EVP}}                     & 0.088 & 0.904 & 0.844 & 0.857 & \multicolumn{1}{c|}{0.891} & 0.033 & 0.841 & 0.817 & 0.886 & \multicolumn{1}{c|}{0.954} & 0.020  & 0.608 & 0.647 & 0.815 & \multicolumn{1}{c|}{0.831} & 0.034  & 0.665  & 0.667  & 0.801 & 0.868 \\ 
			\multicolumn{1}{c|}{FSPNet$_{23}$\cite{FSPNet}}                  & {0.060} & 0.881 & {0.887} & {0.892} & \multicolumn{1}{c|}{0.898} & 0.029 & {0.873} & 0.844 & 0.900 & \multicolumn{1}{c|}{{0.965}} & {0.017} & 0.647 & 0.666 & {0.823} & \multicolumn{1}{c|}{0.867} & 0.030  & 0.690  & 0.678  & 0.814 & 0.886 \\ \hline
            \rowcolor{blue!8}\multicolumn{1}{c|}{Ours-P}                  & \textbf{\color{green}0.051} & \textbf{\color{green}0.925} & \textbf{\color{green}0.902} & \textbf{\color{green}0.893} & \multicolumn{1}{c|}{\textbf{\color{green}0.934}} & \textbf{\color{red}0.022} & \textbf{\color{red}0.891} & \textbf{\color{red}0.887} & \textbf{\color{red}0.914} & \multicolumn{1}{c|}{\textbf{\color{red}0.970}} & \textbf{\color{red}0.015} & \textbf{\color{red}0.738} & \textbf{\color{red}0.745} & \textbf{\color{red}0.850} & \multicolumn{1}{c|}{\textbf{\color{red}0.913}} & \textbf{\color{green}0.028}  & \textbf{\color{green}0.743}  & \textbf{\color{green}0.736}  & \textbf{\color{red}0.829} & \textbf{\color{green}0.911} \\ 
			\rowcolor{blue!8}\multicolumn{1}{c|}{Ours-S}                 & \textbf{\color{red}0.047} & \textbf{\color{red}0.931} & \textbf{\color{red}0.914} & \textbf{\color{red}0.903} & \multicolumn{1}{c|}{\textbf{\color{red}0.937}} & \textbf{\color{green}0.023} & \textbf{\color{green}0.879} & \textbf{\color{green}0.873} & \textbf{\color{green}0.906} & \multicolumn{1}{c|}{\textbf{\color{green}0.966}} & \textbf{\color{green}0.016} & \textbf{\color{green}0.715} & \textbf{\color{green}0.724} & \textbf{\color{green}0.839} & \multicolumn{1}{c|}{\textbf{\color{green}0.906}} & \textbf{\color{red}0.026}  & \textbf{\color{red}0.748}  & \textbf{\color{red}0.738}  & \textbf{\color{red}0.829} & \textbf{\color{red}0.921} \\ \toprule[1pt]\toprule[1pt]
		\end{tabular}}
\label{TABLE-4-1}
\end{table*}

\begin{table*}[]
\renewcommand{\arraystretch}{0.8}
	\setlength{\tabcolsep}{3pt}
	\centering
	\caption{ Quantitative results on the COD10K of other four classes ($i.e.$, ``Aquatic'', ``Terrestrial'', ``Flying'', and ``Amphibian'). The best two results are shown in \textbf{\color{red} red} and \textbf{\color{green} green}. The symbols ``$\uparrow$/$\downarrow$'' denote that a higher/lower value is better.}
	\resizebox*{\textwidth}{70mm}{
\begin{tabular}{ccccccccccccccccccccc}
\toprule[1pt]\toprule[1pt]
\multicolumn{1}{c|}{\multirow{2}{*}{Method}} & \multicolumn{5}{c|}{Aquatic (474 images)}                          & \multicolumn{5}{c|}{Terrestrial (699 images) }                      & \multicolumn{5}{c|}{Flying (714 images) }                           & \multicolumn{5}{c}{Amphibian (124 images) }    \\  
\multicolumn{1}{c|}{}                       & $MAE$ $\downarrow$   & $AF_m$ $\uparrow$   & $WF_m$  $\uparrow$   & $S_m$ $\uparrow$    & \multicolumn{1}{c|}{$E_m$ $\uparrow$}   & $MAE$ $\downarrow$   & $AF_m$ $\uparrow$   & $WF_m$  $\uparrow$   & $S_m$ $\uparrow$    & \multicolumn{1}{c|}{$E_m$ $\uparrow$}   & $MAE$ $\downarrow$   & $AF_m$ $\uparrow$   & $WF_m$  $\uparrow$   & $S_m$ $\uparrow$    & \multicolumn{1}{c|}{$E_m$ $\uparrow$}  & $MAE$ $\downarrow$   & $AF_m$ $\uparrow$   & $WF_m$  $\uparrow$   & $S_m$ $\uparrow$    & \multicolumn{1}{c}{$E_m$ $\uparrow$}   \\ \hline
\multicolumn{21}{c}{Convolution backbone-based COD Method}                                                                                                                                                                                                                                                      \\ \hline
\multicolumn{1}{c|}{SINet$_{20}$\cite{COD10K}}                   & 0.065 & 0.670 & 0.644 & 0.778 & \multicolumn{1}{c|}{0.836} & 0.046 & 0.577 & 0.565 & 0.754 & \multicolumn{1}{c|}{0.810} & 0.040 & 0.614 & 0.580 & 0.797 & \multicolumn{1}{c|}{0.817} & 0.042 & 0.684 & 0.654 & 0.827 & 0.847 \\ 
\rowcolor{blue!8}\multicolumn{1}{c|}{TINet$_{21}$\cite{TINet}}                   & 0.063 & 0.666 & 0.640 & 0.782 & \multicolumn{1}{c|}{0.836} & 0.042 & 0.604 & 0.587 & 0.764 & \multicolumn{1}{c|}{0.832} & 0.032 & 0.679 & 0.668 & 0.817 & \multicolumn{1}{c|}{0.869} & 0.034 & 0.725 & 0.704 & 0.831 & 0.876 \\ 
\multicolumn{1}{c|}{UGTR$_{21}$\cite{UGTR}}                    & 0.050 & 0.707 & 0.687 & 0.808 & \multicolumn{1}{c|}{0.860} & \textbf{\color{green}0.036} & 0.609 & 0.606 & 0.786 & \multicolumn{1}{c|}{0.820} & 0.026 & 0.697 & 0.700 & 0.841 & \multicolumn{1}{c|}{0.870} & 0.029 & 0.749 & 0.738 & \textbf{\color{green}0.857} & 0.883 \\ 
\rowcolor{blue!8}\multicolumn{1}{c|}{JSOCOD$_{21}$\cite{JSOCOD}}                  & 0.049 & 0.734 & 0.705 & 0.803 & \multicolumn{1}{c|}{0.877} & 0.037 & 0.647 & 0.624 & 0.776 & \multicolumn{1}{c|}{0.863} & 0.026 & 0.733 & 0.719 & 0.834 & \multicolumn{1}{c|}{0.903} & 0.031 & 0.765 & 0.742 & 0.841 & 0.903 \\ 
\multicolumn{1}{c|}{MGL-S$_{21}$\cite{MGL}}                   & 0.053 & 0.698 & 0.674 & 0.801 & \multicolumn{1}{c|}{0.858} & 0.037 & 0.606 & 0.593 & 0.778 & \multicolumn{1}{c|}{0.823} & 0.027 & 0.693 & 0.690 & 0.835 & \multicolumn{1}{c|}{0.871} & 0.029 & 0.754 & 0.734 & 0.855 & 0.886 \\ 
\rowcolor{blue!8}\multicolumn{1}{c|}{PFNet$_{21}$\cite{PFNet}}                   & 0.055 & 0.706 & 0.675 & 0.791 & \multicolumn{1}{c|}{0.865} & 0.041 & 0.622 & 0.606 & 0.770 & \multicolumn{1}{c|}{0.850} & 0.030 & 0.698 & 0.691 & 0.822 & \multicolumn{1}{c|}{0.887} & 0.031 & 0.753 & 0.740 & 0.848 & 0.896 \\ 
\multicolumn{1}{c|}{LSR$_{21}$\cite{NC4K}}                     & 0.053 & 0.727 & 0.694 & 0.801 & \multicolumn{1}{c|}{0.875} & 0.038 & 0.639 & 0.611 & 0.770 & \multicolumn{1}{c|}{0.858} & 0.027 & 0.726 & 0.707 & 0.828 & \multicolumn{1}{c|}{0.907} & 0.030 & 0.774 & 0.751 & 0.845 & 0.922 \\ 
\rowcolor{blue!8}\multicolumn{1}{c|}{C2FNet$_{21}$\cite{C2FNet}}                  & 0.052 & 0.727 & 0.701 & 0.805 & \multicolumn{1}{c|}{0.879} & 0.037 & 0.644 & 0.627 & 0.781 & \multicolumn{1}{c|}{0.864} & 0.026 & 0.734 & 0.724 & 0.838 & \multicolumn{1}{c|}{0.908} & 0.030 & 0.771 & 0.752 & 0.849 & 0.919 \\ 
\multicolumn{1}{c|}{CubeNet$_{22}$\cite{CubeNet}}                 & 0.058 & 0.705 & 0.668 & 0.792 & \multicolumn{1}{c|}{0.863} & 0.042 & 0.610 & 0.581 & 0.760 & \multicolumn{1}{c|}{0.838} & 0.031 & 0.698 & 0.677 & 0.817 & \multicolumn{1}{c|}{0.883} & 0.035 & 0.736 & 0.714 & 0.836 & 0.893 \\ 
\rowcolor{blue!8}\multicolumn{1}{c|}{ERRNet$_{22}$\cite{ERRNet}}                  & 0.060 & 0.676 & 0.649 & 0.780 & \multicolumn{1}{c|}{0.847} & 0.045 & 0.580 & 0.565 & 0.751 & \multicolumn{1}{c|}{0.816} & 0.032 & 0.676 & 0.667 & 0.812 & \multicolumn{1}{c|}{0.867} & 0.036 & 0.736 & 0.717 & 0.834 & 0.885 \\ 
\multicolumn{1}{c|}{PreyNet$_{22}$\cite{PreyNet}}                 & 0.049 & \textbf{\color{red}0.757} & \textbf{\color{green}0.718} & 0.810 & \multicolumn{1}{c|}{\textbf{\color{red}0.892}} & \textbf{\color{green}0.036} & 0.664 & 0.628 & 0.774 & \multicolumn{1}{c|}{\textbf{\color{green}0.867}} & \textbf{\color{green}0.024} & \textbf{\color{red}0.770} & \textbf{\color{green}0.740} & 0.840 & \multicolumn{1}{c|}{\textbf{\color{red}0.918}} & \textbf{\color{green}0.028} & \textbf{\color{red}0.797} & 0.768 & 0.856 & \textbf{\color{green}0.927} \\ 
\rowcolor{blue!8}\multicolumn{1}{c|}{SegMaR$_{22}$\cite{SegMaR}}                  & \textbf{\color{green}0.048} & 0.724 & 0.704 & 0.812 & \multicolumn{1}{c|}{0.878} & 0.037 & 0.642 & 0.623 & 0.780 & \multicolumn{1}{c|}{0.861} & 0.027 & 0.725 & 0.713 & 0.834 & \multicolumn{1}{c|}{0.899} & 0.031 & 0.777 & 0.755 & 0.854 & 0.913 \\ 
\multicolumn{1}{c|}{FAPNet$_{22}$\cite{FAPNet}}                  & 0.049 & 0.740 & 0.717 & \textbf{\color{green}0.817} & \multicolumn{1}{c|}{0.879} & 0.037 & 0.652 & 0.639 & \textbf{\color{red}0.792} & \multicolumn{1}{c|}{0.852} & 0.025 & 0.733 & 0.725 & \textbf{\color{green}0.843} & \multicolumn{1}{c|}{0.896} & 0.032 & 0.768 & 0.752 & 0.854 & 0.896 \\ 
\rowcolor{blue!8}\multicolumn{1}{c|}{BSANet$_{22}$\cite{BSANet}}                  & 0.050 & 0.738 & 0.706 & 0.807 & \multicolumn{1}{c|}{0.886} & \textbf{\color{red}0.035} & \textbf{\color{red}0.668} & \textbf{\color{green}0.643} & 0.787 & \multicolumn{1}{c|}{\textbf{\color{red}0.875}} & \textbf{\color{green}0.024} & 0.759 & 0.738 & 0.841 & \multicolumn{1}{c|}{0.913} & \textbf{\color{red}0.027} & 0.786 & \textbf{\color{green}0.769} & \textbf{\color{red}0.858} & \textbf{\color{red}0.931} \\ 
\multicolumn{1}{c|}{MRRNet$_{23}$}\cite{TNNLS3}                  & 0.052 & 0.711 & 0.694 & 0.807 & \multicolumn{1}{c|}{0.863} & {0.037} & {0.635} & {0.633} & \textbf{\color{green}0.790} & \multicolumn{1}{c|}{0.842} & {0.025} & 0.735 & 0.738 & \textbf{\color{red}0.851} & \multicolumn{1}{c|}{0.896} & {0.033} & 0.765 & {0.756} & {0.856} & {0.892} \\
\hline
\rowcolor{blue!8}\multicolumn{1}{c|}{Ours-R}                  & \textbf{\color{red}0.045} & \textbf{\color{green}0.755} & \textbf{\color{red}0.732} & \textbf{\color{red}0.820} & \multicolumn{1}{c|}{\textbf{\color{green}0.891}} & \textbf{\color{red}0.035} & \textbf{\color{green}0.667} & \textbf{\color{red}0.652} & {0.789} & \multicolumn{1}{c|}{0.864} & \textbf{\color{red}0.023} & \textbf{\color{green}0.764} & \textbf{\color{red}0.755} & \textbf{\color{red}0.851} & \multicolumn{1}{c|}{\textbf{\color{green}0.915}} & 0.031 & \textbf{\color{green}0.791} & \textbf{\color{red}0.772} & 0.856 & 0.913 \\ \toprule[1pt]\toprule[1pt]
\multicolumn{21}{c}{Transformer backbone-based COD Method}                                                                                                                                                                                                                                                     \\ \hline
\rowcolor{blue!8}\multicolumn{1}{c|}{VST$_{21}$\cite{VST}}                     & 0.050 & 0.742 & 0.714 & 0.817 & \multicolumn{1}{c|}{0.880} & 0.041 & 0.656 & 0.629 & 0.780 & \multicolumn{1}{c|}{0.855} & 0.025 & 0.757 & 0.739 & 0.845 & \multicolumn{1}{c|}{0.907} & 0.032 & 0.707 & 0.789 & 0.870 & 0.922 \\ 
\multicolumn{1}{c|}{FSPNet$_{23}$\cite{FSPNet}}                  & \textbf{\color{green}0.037} & 0.764 & 0.763 & {0.847} & \multicolumn{1}{c|}{{0.905}} & {\textbf{\color{green}0.027}} & 0.688 & 0.679 & 0.819 & \multicolumn{1}{c|}{0.880} & {\textbf{\color{green}0.020}} & 0.759 & 0.761 & 0.868 & \multicolumn{1}{c|}{0.913} & 0.023 & 0.786 & 0.782 & 0.878 & 0.927 \\ 
\rowcolor{blue!8}\multicolumn{1}{c|}{EVP$_{23}$\cite{EVP}}                     & 0.047 & 0.735 & 0.724 & 0.828 & \multicolumn{1}{c|}{0.875} & 0.033 & 0.645 & 0.654 & 0.804 & \multicolumn{1}{c|}{0.849} & 0.023 & 0.738 & 0.751 & 0.860 & \multicolumn{1}{c|}{0.903} & 0.026 & 0.785 & {0.790} & {0.882} & 0.917 \\ 
\multicolumn{1}{c|}{FPNet$_{23}$\cite{FPNet}}                   & 0.042 & {0.788} & {0.767} & 0.840 & \multicolumn{1}{c|}{0.898} & 0.029 & {0.720} & {0.711} & {0.825} & \multicolumn{1}{c|}{{0.896}} & {\textbf{\color{green}0.020}} & {0.791} & {0.786} & {0.870} & \multicolumn{1}{c|}{{0.926}} & {\textbf{\color{green}0.022}} & {0.800} & 0.783 & 0.871 & {0.942} \\ \hline
\rowcolor{blue!8}\multicolumn{1}{c|}{Ours-P}                   & \textbf{\color{red}0.031} & \textbf{\color{red}0.815} & \textbf{\color{red}0.806} & \textbf{\color{red}0.862} & \multicolumn{1}{c|}{\textbf{\color{green}0.925}} & \textbf{\color{red}0.024} & \textbf{\color{red}0.746} & \textbf{\color{red}0.745} & \textbf{\color{green}0.840} & \multicolumn{1}{c|}{\textbf{\color{red}0.913}} & \textbf{\color{red}0.016} & \textbf{\color{red}0.823} & \textbf{\color{red}0.826} & \textbf{\color{red}0.889} & \multicolumn{1}{c|}{\textbf{\color{red}0.946}} & \textbf{\color{red}0.020} & \textbf{\color{green}0.843} & \textbf{\color{red}0.835} & \textbf{\color{red}0.897} & \textbf{\color{red}0.954} \\ 
\rowcolor{blue!8}\multicolumn{1}{c|}{Ours-S}                  & \textbf{\color{red}0.031} & \textbf{\color{green}0.805} & \textbf{\color{green}0.796} & \textbf{\color{green}0.857} & \multicolumn{1}{c|}{\textbf{\color{red}0.928}} & \textbf{\color{red}0.024} & \textbf{\color{green}0.736} & \textbf{\color{green}0.735} & \textbf{\color{red}0.835} & \multicolumn{1}{c|}{\textbf{\color{green}0.910}} & \textbf{\color{red}0.016} & \textbf{\color{green}0.809} & \textbf{\color{green}0.813} & \textbf{\color{green}0.882} & \multicolumn{1}{c|}{\textbf{\color{green}0.945}} & \textbf{\color{red}0.020} & \textbf{\color{red}0.846} & \textbf{\color{green}0.831} & \textbf{\color{green}0.891} & \textbf{\color{green}0.953} \\ \toprule[1pt]\toprule[1pt]
\end{tabular}}
\label{TABLE-4-2}
\end{table*}
\subsection{Experimental Settings}
\subsubsection{Datasets}
We evaluate our GLCONet method on three widely-used COD datasets, including CAMO \cite{CAMO}, COD10K \cite{COD10K}, and NC4K \cite{NC4K}. CAMO \cite{CAMO} contains 1250 images with camouflaged objects, including 1000 training images and 250 testing images. COD10K \cite{COD10K} is a large-scale dataset with abundant classes of camouflaged objects, which can be divided into a training set of 3040 images and a testing set of 2026 images. NC4K \cite{NC4K} is another large dataset, which contains 4121 testing images. The proposed GLCONet model trains on CAMO-Train \cite{CAMO} dataset of 1000 images and COD10K-Train \cite{COD10K} dataset of 3040 images.
\subsubsection{Implementation details}
We implement GLCONet in the PyTorch framework and train on four NVIDIA GTX 3090 GPUs. We utilize pre-trained ResNet-50 \cite{ResNet}, Swin Transformer \cite{Swin}, and Pyramid Vision Transformer \cite{PVT} as an independent encoder, and use Adam with an initial learning rate of 1e-4 as the optimizer and decay rate of 0.1 for every 60 epochs. Batch size is set to 36 during training and the whole network is iterated 180 epochs. Note that all input images are resized as 384$\times$384 during the training and testing process. Additionally, similar to SINet \cite{COD10K}, BSANet \cite{BSANet}, and FSPNet \cite{FSPNet}, we employ several data augmentation strategies ($e.g.$, horizontal flipping, random cropping, and color enhancement) during the training process to prevent model overfitting.
\subsubsection{Evaluation Metrics}
We exploit seven widely-used evaluation metrics, including mean absolute error ($MAE$), average F-measure ($AF_m$), weight F-measure ($WF_m$), S-measure ($S_m$)\nocite{Sm}, E-measure ($E_m$), Precision-Recall ($PR$) curve, and F-measure ($F_m$) curve.

\textbf{Mean absolute error ($MAE$)} is applied to measure the pixel-wise difference between the predicted map ($P_{ij}$) and ground truth ($G_{ij}$), which is defined as follows: 
\begin{equation}   
	\begin{aligned} 
		MAE\;=\;\frac1{W\times H}\sum_{i=1\;}^W\sum_{j=1}^H\vert P_{ij}-G_{ij}\vert,
	\end{aligned}
\end{equation}  
where $H$ and $W$ denote the height and width of the ground truth, respectively; $(i,j)$ is the pixel's position.

\textbf{F-measure ($F_m$)} is a comprehensive evaluation metric, which is adopted as the weighted harmonic mean of precisions and recalls, $F_m$ can be formulated as follows:
\begin{equation}   
	\begin{aligned} 
		\mbox{\large $F_m$}\,=\frac{(1+\beta^2)\times Precision\times Recall}{\beta^2\times Precision + Recall},
	\end{aligned}
\end{equation}
where $\beta^2$ = 0.3 to highlight the precision over recall. We utilize the average and weight values ($i.e.$, $AF_m$ and $WF_m$) \cite{AFm} computed from all precision-recall pairs in this paper.

\textbf{S-measure ($S_m$) \cite{Sm}} is a standard metric to assess the structural similarity at the region ($S_r$) and object ($S_o$) level, that is,
\begin{equation}   
	\begin{aligned} 
		\mbox{\large$S_m$}\,\;=\;\alpha\mbox{\large$S_r$}\,-(1-\alpha)\mbox{\large$S_o$}\,,
	\end{aligned}
\end{equation}  
where $\alpha$ is a trade-off coefficient, which is set to 0.5.

\textbf{E-measure ($E_m$)} estimates the similarity between the predicted map ($P_{ij}$) and ground truth ($G_{ij}$) by considering local and global similarities and combining local pixel scores with image-wise averages, which is defined as follows:
\begin{equation}   
	\begin{aligned} 
		E_m\;=\;\frac1{W\times H}\sum_{i=1}^W\sum_{j=1}^H\mathcal{Z}(P_{ij},G_{ij}),
	\end{aligned}
\end{equation}
where $\mathcal{Z}(\cdot, \cdot)$ denotes the enhanced alignment matrix, and $(i,j)$ is the pixel's position.

\begin{table*}[t]
	\renewcommand{\arraystretch}{1}
	\setlength{\tabcolsep}{2.5pt}
	\centering
	\caption{Ablation analysis of our GLCONet structure. The best results are shown in \textbf{\color{red} red}.}
	\resizebox{0.88\textwidth}{20mm}{
		\begin{tabular}{cccccc|ccccc|ccccc}
			\toprule[1pt]\toprule[1pt]
			\multicolumn{6}{c|}{\normalsize Structure settings}                                                                                                                                                                             & \multicolumn{5}{c|}{\multirow{2}{*}{\normalsize CAMO (250 images)}} & \multicolumn{5}{c}{\multirow{2}{*}{\normalsize COD10K (2026 images)}} \\ \cline{1-6}
			\multicolumn{1}{c|}{\multirow{2}{*}{\begin{tabular}[c]{@{}c@{}} Baseline\end{tabular}}} & \multicolumn{3}{c|}{COS}              & \multicolumn{1}{c|}{\multirow{2}{*}{MTB}} & \multirow{2}{*}{ARD} & \multicolumn{5}{c|}{}                                  & \multicolumn{5}{c}{}                                     \\
			\multicolumn{1}{c|}{}                                                                             & GPM & LRM & \multicolumn{1}{c|}{GHIM} & \multicolumn{1}{c|}{}                     &                      & $MAE$ $\downarrow$   & $AF_m$ $\uparrow$   & $WF_m$ $\uparrow$   & $S_m$ $\uparrow$    & $E_m$ $\uparrow$   & $MAE$ $\downarrow$  & $AF_m$ $\uparrow$  & $WF_m$ $\uparrow$  & $S_m$ $\uparrow$   & $E_m$ $\uparrow$       \\ \hline
			\multicolumn{1}{c|}{$\checkmark$}                                                                            &     &     & \multicolumn{1}{c|}{}     & \multicolumn{1}{c|}{}                     &                      & 0.159     & 0.530     & 0.400    & 0.605    & 0.753    & 0.080     & 0.474     & 0.375     & 0.653     & 0.752    \\ 
			\rowcolor{blue!8}\multicolumn{1}{c|}{$\checkmark$}                                                                            & $\checkmark$   &     & \multicolumn{1}{c|}{}     & \multicolumn{1}{c|}{}                     &                      & 0.076     & 0.761     & 0.710    & 0.791    & 0.862    & 0.036     & 0.710     & 0.691     & 0.809     & 0.883    \\ 
			\multicolumn{1}{c|}{$\checkmark$}                                                                            &     & $\checkmark$   & \multicolumn{1}{c|}{}     & \multicolumn{1}{c|}{}                     &                      & 0.078     & 0.761     & 0.714    & 0.801    & 0.868    & 0.039     & 0.681     & 0.661     & 0.801     & 0.863    \\ 
			\rowcolor{blue!8}\multicolumn{1}{c|}{$\checkmark$}                                                                            & $\checkmark$   & $\checkmark$   & \multicolumn{1}{c|}{}     & \multicolumn{1}{c|}{}                     &                      & 0.075     & 0.762     & 0.719    & 0.800    & 0.869    & 0.035     & 0.695     & 0.692     & 0.815     & 0.872    \\ 
			\multicolumn{1}{c|}{$\checkmark$}                                                                            & $\checkmark$   & $\checkmark$   & \multicolumn{1}{c|}{$\checkmark$}    & \multicolumn{1}{c|}{}                     &                      & 0.072     & 0.771     & 0.734    & 0.810    & 0.879    & 0.036     & 0.708     & 0.695     & 0.814     & 0.878    \\ 
			\rowcolor{blue!8}\multicolumn{1}{c|}{$\checkmark$}                                                                            &     &     & \multicolumn{1}{c|}{}     & \multicolumn{1}{c|}{$\checkmark$}                    &                      & 0.147     & 0.563     & 0.438    & 0.615    & 0.763    & 0.069     & 0.515     & 0.429     & 0.663     & 0.780    \\ 
			\multicolumn{1}{c|}{$\checkmark$}                                                                            &     &     & \multicolumn{1}{c|}{}     & \multicolumn{1}{c|}{$\checkmark$}                    & $\checkmark$                    & 0.076     & 0.766     & 0.710    & 0.794    & 0.869    & 0.035     & 0.708     & 0.691     & 0.812     & 0.882    \\ 
			\rowcolor{blue!8} \multicolumn{1}{c|}{$\checkmark$}                                                                            & $\checkmark$   & $\checkmark$   & \multicolumn{1}{c|}{$\checkmark$}    & \multicolumn{1}{c|}{$\checkmark$}                    & $\checkmark$                    & \textbf{\color{red} 0.069}     & \textbf{\color{red} 0.788}     & \textbf{\color{red} 0.748}    & \textbf{\color{red} 0.816}    & \textbf{\color{red} 0.882}    & \textbf{\color{red} 0.033}     & \textbf{\color{red} 0.729}     & \textbf{\color{red} 0.714}     & \textbf{\color{red} 0.822}     & \textbf{\color{red} 0.891}    \\ \toprule[1pt]\toprule[1pt]
		\end{tabular}}
\label{TABLE-A-1}
\end{table*}

\subsection{Comparisons with the State-of-the-Art Methods}
We compare our GLCONet model with multiple SOTA methods, including SINet \cite{COD10K}, PFNet \cite{PFNet}, MGL-S \cite{MGL}, LSR \cite{NC4K}, JSOCOD \cite{JSOCOD}, UGTR \cite{UGTR}, C2FNet \cite{C2FNet}, TINet \cite{TINet}, ERRNet \cite{ERRNet}, CubeNet \cite{CubeNet}, PreyNet \cite{PreyNet}, FAPNet \cite{FAPNet}, BSANet \cite{BSANet}, SegMaR \cite{SegMaR}, PUENet \cite{PUENet}, MRRNet \cite{TNNLS3}, VST \cite{VST}, FPNet \cite{FPNet}, EVP \cite{EVP}, and FSPNet \cite{FSPNet}. For a fair comparison, we distinguish between Transformer backbone-based methods and Convolution backbone-based methods. In addition, all prediction maps in this paper are provided by the authors or obtained from open-source codes.

\subsubsection{Quantitative comparisons}
Table \ref{TABLE2} summarizes the quantitative result of our GLCONet against the 20 SOTA methods. From Table \ref{TABLE2}, it can be seen that the proposed GLCONet method achieves excellent performance regardless of the type of backbone network it is based on. Specifically, compared with the second best method, the GLCONet method with ResNet-50 \cite{ResNet} backbone shows declined $MAE$, it is reduced by 4.35\%, 3.03\%, and 9.30\% on three public COD datasets. Moreover, compared to the recent MRRNet \cite{TNNLS3} method, there are significant improvements in $AF_m$ and $WF_m$ metrics, with increases of 2.87\% and 2.75\% on CAMO \cite{CAMO} dataset, 4.89\% and 3.18\% on COD10K \cite{COD10K} dataset, and 3.55\% and 3.26\% on NC4K \cite{NC4K} dataset, respectively.  Similarly, in terms of other evaluation metrics, our GLCONet method achieves superior performance. Furthermore, compared to the Transformer backbone-based methods ($e.g.$, FSPNet \cite{FSPNet}), the Swin Transformer version of our approach attains a remarkable performance boost of 31.57\%, 4.22\%, 6.26\%, 2.92\%, and 2.29\% on CAMO \cite{CAMO} dataset, 13.04\%, 6.67\%, 6.53\%, 1.53\%, and 3.22\% on COD10K \cite{COD10K} dataset, and 16.67\%, 3.87\%, 3.80\%, 0.91\%, and 2.06\% on NC4K \cite{NC4K} dataset in terms of $MAE$, $AF_m$, $WF_m$, $E_m$, and $S_m$. Besides, with the same Pyramid Vision Transformer \cite{PVT} backbone, the proposed GLCONet method shows significant advantages over the current FPNet \cite{FPNet} model. For example, on two datasets ($i.e.$, CAMO \cite{CAMO} dataset, and COD10K \cite{COD10K} dataset), the proposed GLCONet method improves by 5.61\% and 5.16\% on the $WF_m$ metric, and by 3.17\% and 2.24\% on the $S_m$ metric, respectively. Similar performance improvements are also achieved on other evaluation metrics. In addition, in Fig. \ref{Fig.6}, we present the $PR$ and $F_m$ curves, with our GLCONet method indicated in red line. It can be observed that compared to existing COD methods, the GLOCNet model has significant advantages. These above results demonstrate the effectiveness of our GLCONet method in detecting camouflaged objects across diverse scenarios, which is due to the joint optimization of local details and global relationships for input features achieved by the designed collaborative optimization strategy. 
\begin{figure}[]
		\centering\includegraphics[width=0.48\textwidth,height=3.8cm]{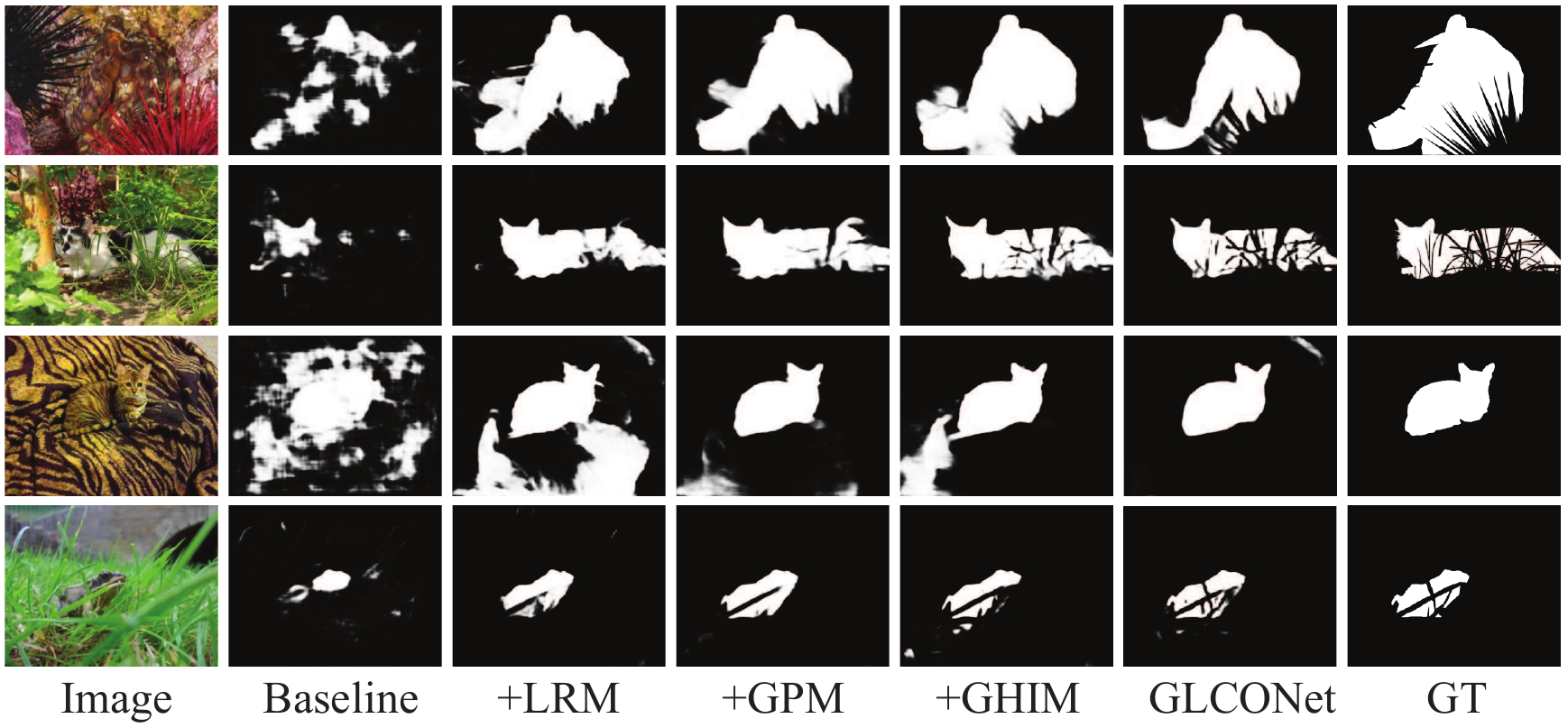}
		\captionsetup{font={small}, justification=raggedright}
		\caption{ Visual results of the effectiveness of our modules.}
		\label{Fig.8}
\end{figure}
\subsubsection{Qualitative comparisons}
Fig. \ref{Fig.7} shows the visual comparison of our GLCONet method with existing COD methods in several scenes, including large size ($1^{st}$- $3^{rd}$ rows), small size ($4^{th}$- $6^{th}$ rows), and occluded objects ($7^{th}$- $9^{th}$ rows). As depicted in the first six rows of Fig. \ref{Fig.7}, some SOTA competitors ($e.g.$, SegMaR \cite{SegMaR}, BSANet \cite{BSANet}, and FAPNet \cite{FAPNet}) are unable to completely segment camouflaged objects with unfixed size. In contrast, thanks to the inter-collaborative optimization of the designed module, our GLCONet method is sensitive to camouflaged objects of different sizes, which can efficiently segment objects from different scenarios. In addition, from $7^{th}$- $9^{th}$ rows of Fig. \ref{Fig.7}, it can be seen that compared to existing SOTA methods, the proposed GLCONet method has great accuracy for occluded objects.
\begin{figure}[]
 	\centering\includegraphics[width=0.48\textwidth,height=3.8cm]{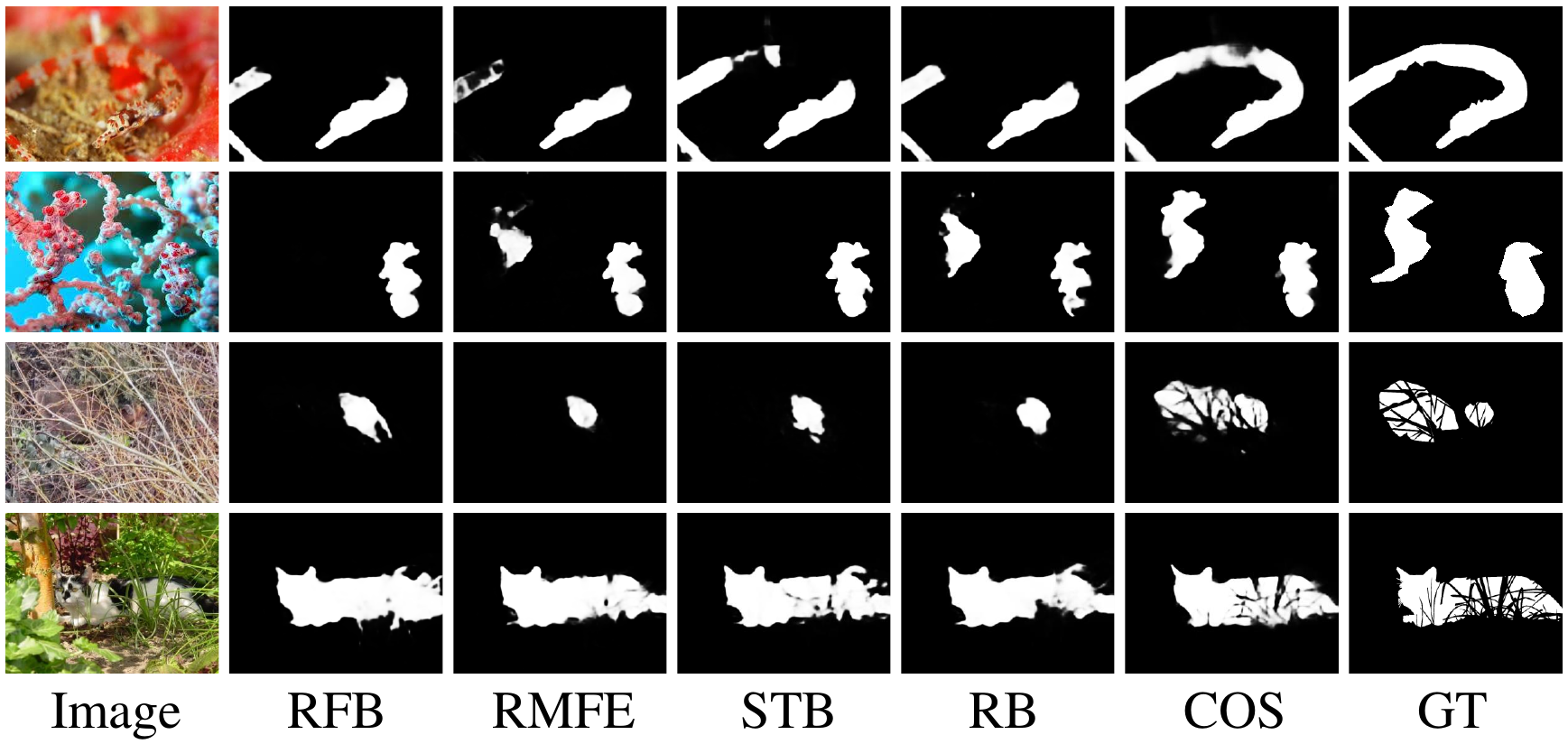}
 	\captionsetup{font={small}, justification=raggedright}
 	\caption{ Visual results between our COS and other feature enhancement modules. ``GT'' denotes the ground truth.}
 	\label{Fig.9}
\end{figure}
\begin{table}[t]
 \renewcommand{\arraystretch}{1}
 \setlength{\tabcolsep}{2pt}
 \centering
 \caption{Ablation analysis of our COS $vs.$ existing feature enhancement modules.}
 	\resizebox{0.5\textwidth}{11mm}{
 		\begin{tabular}{l|ccccc|ccccc}
 			\toprule[1pt]\toprule[1pt]
 			\multicolumn{1}{c|}{\multirow{2}{*}{\normalsize Components}} & \multicolumn{5}{c|}{CAMO-Test(250 images)}   & \multicolumn{5}{c}{COD10K-Test (2026 images)} \\
 			\multicolumn{1}{c|}{}                           & $MAE$ $\downarrow$   & $AF_m$ $\uparrow$   & $WF_m$ $\uparrow$   & $S_m$ $\uparrow$    & $E_m$ $\uparrow$   & $MAE$ $\downarrow$  & $AF_m$ $\uparrow$  & $WF_m$ $\uparrow$  & $S_m$ $\uparrow$   & $E_m$ $\uparrow$    \\ \hline 
 			\rowcolor{blue!8}Baseline + RFB \cite{RFB}                                      & 0.079 & 0.756 & 0.711 & 0.789 & 0.867 & 0.037 & 0.694 & 0.673 & 0.797 & 0.879 \\ 
 			Baseline + RMEF \cite{BSANet}                                      & 0.079 & 0.755 & 0.706 & 0.784 & 0.863 & 0.035 & 0.707 & 0.687 & 0.803 & 0.884 \\ 
 			\rowcolor{blue!8}Baseline + STB \cite{Swin}                                     & 0.077 & 0.757 & 0.721 & 0.796 & 0.865 & 0.039 & 0.690 & 0.679 & 0.804 & 0.871 \\ 
 			Baseline + RB \cite{Restormer}                                      & 0.079 & 0.749 & 0.698 & 0.778 & 0.859 & 0.036 & 0.701 & 0.678 & 0.799 & 0.884 \\ 
 			\rowcolor{blue!8}Baseline + COS                                       & \textbf{\color{red} 0.072} & \textbf{\color{red} 0.777} & \textbf{\color{red} 0.735} & \textbf{\color{red} 0.807} & \textbf{\color{red} 0.878} & \textbf{\color{red} 0.034} & \textbf{\color{red} 0.724} & \textbf{\color{red} 0.706} & \textbf{\color{red} 0.818} & \textbf{\color{red} 0.893} \\ \toprule[1pt]\toprule[1pt]
 		\end{tabular}}
        \label{TABLE-A-COS-1}
 	\end{table}
\begin{table*}[]
\renewcommand{\arraystretch}{1}
	\setlength{\tabcolsep}{3pt}
    \centering
	\caption{Ablation analysis of the MTB and PCB in multi-scale spaces.}
	\resizebox{0.88\textwidth}{27mm}{
\begin{tabular}{cccccccccccccc}
\toprule[1pt]\toprule[1pt]
\multicolumn{1}{c|}{\multirow{2}{*}{Baseline}} & \multicolumn{3}{c|}{MTB}                         & \multicolumn{5}{c|}{CAMO (250 images)}                     & \multicolumn{5}{c}{COD10K (2026 images)} \\  
\multicolumn{1}{c|}{}                          & DC$_3$     & DC$_5$     & \multicolumn{1}{c|}{DC$_7$}     & $MAE$ $\downarrow$   & $AF_m$ $\uparrow$   & $WF_m$ $\uparrow$   & $S_m$ $\uparrow$       & \multicolumn{1}{c|}{$E_m$ $\uparrow$}    & $MAE$ $\downarrow$  & $AF_m$ $\uparrow$  & $WF_m$ $\uparrow$  & $S_m$ $\uparrow$   & $E_m$ $\uparrow$    \\ \hline
\multicolumn{1}{c|}{$\checkmark$}                         & $\checkmark$       &         & \multicolumn{1}{c|}{}        & \textbf{\color{red} 0.076} & 0.754 & 0.708 & 0.785 & \multicolumn{1}{c|}{\textbf{\color{red} 0.866}} & 0.037  & 0.706  & 0.683  & 0.802 & \textbf{\color{red} 0.883} \\ 
\rowcolor{blue!8}\multicolumn{1}{c|}{$\checkmark$}                         &         & $\checkmark$       & \multicolumn{1}{c|}{}        & 0.084 & 0.743 & 0.688 & 0.690 & \multicolumn{1}{c|}{0.852} & 0.037  & 0.700  & 0.671  & 0.792 & \textbf{\color{red} 0.883} \\ 
\multicolumn{1}{c|}{$\checkmark$}                         &         &         & \multicolumn{1}{c|}{$\checkmark$}       & 0.084 & 0.743 & 0.685 & 0.767 & \multicolumn{1}{c|}{0.839} & \textbf{\color{red} 0.036}  & 0.706  & 0.677  & 0.797 & 0.882 \\ 
\rowcolor{blue!8}\multicolumn{1}{c|}{$\checkmark$}                         & $\checkmark$       & $\checkmark$       & \multicolumn{1}{c|}{$\checkmark$}       & \textbf{\color{red} 0.076} & \textbf{\color{red} 0.761} & \textbf{\color{red} 0.710} & \textbf{\color{red} 0.791} & \multicolumn{1}{c|}{0.862} & \textbf{\color{red} 0.036}  & \textbf{\color{red} 0.710}  & \textbf{\color{red} 0.691}  & \textbf{\color{red} 0.809} & \textbf{\color{red} 0.883} \\ \toprule[1pt]\toprule[1pt]
\multicolumn{14}{c}{}                                                                                                                                                                                     \\ \toprule[1pt]\toprule[1pt]
\multicolumn{1}{c|}{\multirow{2}{*}{Baseline}} & \multicolumn{3}{c|}{PCB}                         & \multicolumn{5}{c|}{CAMO (250 images)}                     & \multicolumn{5}{c}{COD10K (2026 images)} \\ 
\multicolumn{1}{c|}{}                          & AC$_3$+DC$_3$ & AC$_5$+DC$_5$ & \multicolumn{1}{c|}{AC$_7$+DC$_7$} & $MAE$ $\downarrow$   & $AF_m$ $\uparrow$   & $WF_m$ $\uparrow$   & $S_m$ $\uparrow$       & \multicolumn{1}{c|}{$E_m$ $\uparrow$}    & $MAE$ $\downarrow$  & $AF_m$ $\uparrow$  & $WF_m$ $\uparrow$  & $S_m$ $\uparrow$   & $E_m$ $\uparrow$    \\ \hline
\multicolumn{1}{c|}{$\checkmark$}                         & $\checkmark$       &         & \multicolumn{1}{c|}{}        & 0.083 & 0.746 & 0.699 & 0.786 & \multicolumn{1}{c|}{0.861} & \textbf{\color{red}0.039}  & 0.668  & \textbf{\color{red} 0.661}  & 0.800 & 0.857 \\ 
\rowcolor{blue!8}\multicolumn{1}{c|}{$\checkmark$}                         &         & $\checkmark$       & \multicolumn{1}{c|}{}        & 0.082 & 0.747 & 0.699 & 0.788 & \multicolumn{1}{c|}{0.864} & 0.040  & 0.669  & 0.654  & 0.799 & 0.859 \\ 
\multicolumn{1}{c|}{$\checkmark$}                         &         &         & \multicolumn{1}{c|}{$\checkmark$}       & 0.081 & 0.741 & 0.704 & 0.793 & \multicolumn{1}{c|}{0.858} & 0.041  & 0.662  & 0.653  & 0.796 & 0.852 \\ 
\rowcolor{blue!8}\multicolumn{1}{c|}{$\checkmark$}                         & $\checkmark$       & $\checkmark$       & \multicolumn{1}{c|}{$\checkmark$}       & \textbf{\color{red} 0.078} & \textbf{\color{red} 0.761} & \textbf{\color{red} 0.714} & \textbf{\color{red} 0.801} & \multicolumn{1}{c|}{\textbf{\color{red} 0.868}} & \textbf{\color{red} 0.039}  & \textbf{\color{red} 0.681}  & \textbf{\color{red} 0.661}  & \textbf{\color{red} 0.801} & \textbf{\color{red} 0.863} \\ \toprule[1pt]\toprule[1pt]
\end{tabular}}
\label{TABLE-A-COS-2}
\end{table*}

\begin{table}[h]
	\centering
	\caption{Ablation analysis of our GHIM structure.}
	\resizebox{0.5\textwidth}{8mm}{
\begin{tabular}{ccc|ccccc|ccccc}
\toprule[1pt]\toprule[1pt]
\multicolumn{3}{c|}{Structure settings}                                   & \multicolumn{5}{c|}{\multirow{2}{*}{CAMO (250 images)}} & \multicolumn{5}{c}{\multirow{2}{*}{COD10K (2026 images)}} \\ \cline{1-3}
\multicolumn{1}{c|}{\multirow{2}{*}{Baseline}} & \multicolumn{2}{c|}{GHIM} & \multicolumn{5}{c|}{}                                   & \multicolumn{5}{c}{}                                      \\ \cline{2-3}
\multicolumn{1}{c|}{}                          & Add         & Cat         & $MAE$ $\downarrow$   & $AF_m$ $\uparrow$   & $WF_m$ $\uparrow$   & $S_m$ $\uparrow$    & $E_m$ $\uparrow$   & $MAE$ $\downarrow$  & $AF_m$ $\uparrow$  & $WF_m$ $\uparrow$  & $S_m$ $\uparrow$   & $E_m$ $\uparrow$       \\ \hline
\multicolumn{1}{c|}{$\checkmark$}                         &             & $\checkmark$           & \textbf{\color{red} 0.072}     & \textbf{\color{red} 0.776}     & \textbf{\color{red} 0.735}     & 0.806    & \textbf{\color{red} 0.882}    & \textbf{\color{red} 0.035}     & 0.702     & 0.694     & 0.812     & \textbf{\color{red} 0.879}    \\ 
\rowcolor{blue!8}\multicolumn{1}{c|}{$\checkmark$}                         & $\checkmark$           &             & \textbf{\color{red} 0.072}     & 0.771     & 0.734     & \textbf{\color{red} 0.810}    & 0.879    & 0.036     & \textbf{\color{red} 0.708}     & \textbf{\color{red} 0.695}     & \textbf{\color{red} 0.814}     & 0.878     \\ \toprule[1pt]\toprule[1pt]
\end{tabular}}
\label{TABLE-A-COS-3}
\end{table}

\subsubsection{More Comparisons}
For a more comprehensive analysis, we divide the COD10K\cite{COD10K} dataset into four categories ($i.e.,$ large objects (165 images), medium objects (543 images), small objects (853 images), and occluded objects (465 images)) to demonstrate the effectiveness of our GLCONet method for camouflaged objects at different sizes. As shown in Table \ref{TABLE-4-1}, the proposed GLCONet method achieves significant advantages against existing COD methods for camouflaged objects at different sizes, especially for large and medium objects with remarkably improved accuracy. Specifically, compared to the second best method, the $MAE$, metric of our GLCONet method with ResNet-50 backbone \cite{ResNet} improves by 1.31\% and 9.68\% on the large and medium object datasets, respectively. Furthermore, by comparision of ``Ours-S'' and ``FSPNet'' \cite{FSPNet}, our GLCONet method remarkably improves by 27.66\%, 5.68\%, 3.04\%, 1.23\% and 4.34\% on large object dataset, 6.25\%, 10.51\%, 8.71\%, 1.94\%, and 4.50\% on small objects dataset, 15.38\%, 8.41\%, 8.85\%, 1.84\%, and 3.95\% on occluded objects dataset for $MAE$, $AF_m$, $WF_m$, $S_m$ and $E_m$. Moreover, the GLCONet method based on feature pyramid network \cite{PVT} achieves significantly excellent performance. Additionally, in Table \ref{TABLE-4-2}, we provide another classification strategy for the COD10K dataset, dividing it into categories ``aquatic (474 images)'', ``terrestrial (699 images)'', ``flying (714 images)'', and ``amphibian (124 images)'' based on different survival environments. It can be seen that our GLCONet method performs significantly well under these classification strategies regardless of which backbone it is based on. Particularly, under the same PVT \cite{PVT} backbone, our GLCONet method outperforms the recent FPNet \cite{FPNet} model in terms of four metrics ($i.e.$, $AF_m$, $WF_m$, $S_m$, and $E_m$), making in improvements of 3.61\%, 4.78\%, 1.82\%, and 1.90\%, 4.05\%, 5.09\%, 2.18\%, and 2.16\% on the amphibian and terrestrial datasets, respectively. This benefits from the proposed collaborative optimization strategy (COS) modeling both global long-range dependencies and local details as well as the cross-layer aggregation and reverse optimization of our adjacent reverse decoder (ARD). 

\begin{table}[t]
	\renewcommand{\arraystretch}{1}
	\setlength{\tabcolsep}{2pt}
	\centering
	\caption{ Ablation analysis of hyper-parameter settings.}
	\resizebox{0.5\textwidth}{12mm}{
		\begin{tabular}{c|cc|c|c|ccccc|ccccc}
			\toprule[1pt]\toprule[1pt]
			\multirow{2}{*}{\normalsize Component} & \multicolumn{2}{c|}{\begin{tabular}[c]{@{}c@{}}Hyper-parameter\end{tabular}} & \multirow{2}{*}{\begin{tabular}[c]{@{}c@{}}Parameters\\ (M)\end{tabular}} & \multirow{2}{*}{\begin{tabular}[c]{@{}c@{}}Flops\\ (G)\end{tabular}} & \multicolumn{5}{c|}{CAMO (250 images)} & \multicolumn{5}{c}{COD10K (2026 images)} \\
			& channel                                 & input size                                 &                                                                         &                                                                      & $MAE$ $\downarrow$   & $AF_m$ $\uparrow$   & $WF_m$ $\uparrow$   & $S_m$ $\uparrow$    & $E_m$ $\uparrow$   & $MAE$ $\downarrow$  & $AF_m$ $\uparrow$  & $WF_m$ $\uparrow$  & $S_m$ $\uparrow$   & $E_m$ $\uparrow$    \\ \hline 
			\rowcolor{blue!8}GLCONet                    & 64                                      & 384$\times$384                                    & \textbf{\color{red} 41.29}                                                                   & \textbf{\color{red} 52.10}                                                                & 0.070  & 0.775 & 0.737 & 0.810 & 0.882 & 0.034  & 0.706  & 0.699  & 0.817 & 0.880 \\ 
			GLCONet                    & 128                                     & 384$\times$384                                    & 70.25                                                                   & 147.68                                                               & 0.069  & 0.787 & 0.748 & 0.816 & 0.882 & 0.033  & 0.729  & 0.714  & 0.822 & 0.891 \\ 
			\rowcolor{blue!8}GLCONet                    & 192                                     & 384$\times$384                                    & 110.40                                                                  & 298.88                                                               & 0.069  & 0.775 & 0.736 & 0.810 & 0.879 & 0.033  & 0.723  & 0.710  & 0.822 & 0.890 \\ 
			GLCONet                    & 128                                     & 352$\times$352                                    & 70.25                                                                   & 124.09                                                               & 0.068  & 0.784 & 0.748 & 0.815 & 0.885 & 0.034  & 0.714  & 0.703  & 0.817 & 0.886 \\ 
			\rowcolor{blue!8}GLCONet                    & 128                                     & 416$\times$416                                    & 70.25                                                                   & 173.32                                                                     & \textbf{\color{red} 0.067}      &  0.789     & \textbf{\color{red} 0.756}      & \textbf{\color{red}0.820}      &  0.885     & 0.032       & 0.735      & 0.725       & 0.828      & 0.896      \\ 
			GLCONet                    & 128                                     & 448$\times$448                                    & 70.25                                                                   & 201.01                                                               & \textbf{\color{red} 0.067}  & \textbf{\color{red} 0.795} & 0.750 &{0.815} & \textbf{\color{red} 0.888} & 0.031  & 0.742  & 0.724  & 0.827 & 0.900 \\ 
			\rowcolor{blue!8}GLCONet                    & 128                                     & 512$\times$512                                    & 70.25                                                                   & 262.55                                                               & 0.069  & 0.787 & 0.741 & 0.809 & 0.879 & \textbf{\color{red} 0.030}  & \textbf{\color{red} 0.753}  & \textbf{\color{red} 0.734}  & \textbf{\color{red} 0.832} & \textbf{\color{red} 0.902} \\ \toprule[1pt]\toprule[1pt]
\end{tabular}}
\label{TABLE-A-H}
\end{table}

\subsection{Ablation Study}
We conduct ablation experiments on two COD datasets ($i.e.$, CAMO \cite{CAMO} and COD10K \cite{COD10K}) under the ResNet-50 \cite{ResNet} backbone. Specifically, we first verify the effectiveness of the global perception module (GPM), the local refinement module (LRM), and the group-wise hybrid interaction module (GHIM) in the collaborative optimization strategy (COS). Then we validate the effectiveness of the adjacent reverse decoder (ARD). Finally, we research some hyper-parameter settings in the proposed GLCONet network.

\subsubsection{Effectiveness of COS}
Collaborative optimization strategy consists of three components, $i.e.$, GPM, LRM, and CHIM, which are designed to adaptively excavate and exploit global long-range relationships and local spatial details for better detecting camouflaged objects. From Table \ref{TABLE-A-1}, the baseline with only ``GPM'' ($2^{nd}$ row) or ``LRM'' ($3^{rd}$ row) has significant advantage compared with ``Baseline'' ($1^{st}$ row). Subsequently, the predicted accuracy is further improved through the merger of the ``LRM'' and ``GPM'' ($4^{th}$ row) and the incorporation of the ``GHIM'' ($5^{th}$ row). The difference between $4^{th}$ row and $5^{th}$ row is whether ``GHIM'' is being used to integrate with global-local information, whereas in ``+LRM+GPM'' ($4^{th}$ row) utilizing simple concatenation to aggregate different information. Moreover, we give out the predicted map of different modules in Fig. \ref{Fig.8}. It can be seen that by continuously incorporating our modules, the model predicts results that are progressively closer to the ground truth (GT). Moreover, we provide some results of the current feature enhancement modules in Table \ref{TABLE-A-COS-1} and Fig. \ref{Fig.9}, including receptive field block (RFB) \cite{RFB}, residual multi-scale feature extractor (RMFE) \cite{BSANet}, Swin Transformer block (STB) \cite{Swin}, and Restormer block (RB) \cite{Restormer}. This demonstrates the significant advantages of our COS over single local or global optimization methods, thanks to the mutual cooperation and joint optimization of both local and global features in the proposed COS. Furthermore, we show the performance of MTB and PCB at different scale spaces in Table \ref{TABLE-A-COS-2}, and it can be seen that it is beneficial to improve the detection accuracy by integrating multi-scale space information. Additionally, we explore the fusion strategy ($i.e.$, element-wise addition and concatenation) used in GHIM, as presented in Table \ref{TABLE-A-COS-3}. It can be seen that regardless of the fusion strategy employed, GHIM demonstrates excellent performance. Based on the above results, it strongly proves the efficiency and rationality of the designed COS in COD tasks, which benefits from the collaborative work of the GPM, LRM, and GHIM.
\begin{table*}[t]
\renewcommand{\arraystretch}{0.9}
	\setlength{\tabcolsep}{2pt}
	\centering
    \caption{Efficiency analysis of our GLCONet model and existing COD methods.}
	\resizebox*{0.90\textwidth}{15mm}{
\begin{tabular}{c|cccccccccccc}
\toprule[1pt]\toprule[1pt]
         & SINet \cite{COD10K} & PFNet\cite{PFNet} & MGL\cite{MGL}    & UGTR\cite{UGTR}    & JSOCOD\cite{JSOCOD} & C2FNet\cite{C2FNet} & SegMaR\cite{SegMaR} & FSPNet\cite{FSPNet} & Ours-R & Ours-S & Ours-P \\ \hline
Paramers (M) & 48.95 & 46.50 & 63.60  & 48.87   & 217.98 & \textbf{\color{red} 28.41}  & 55.62  & 273.79  & 70.25  & 121.59 & 91.35   \\
\rowcolor{blue!8} FLOPs (G)    & 38.75 & 53.22 & 553.94 & 1000.01 & 112.34 & \textbf{\color{red} 26.17}  & 33.65  & 283.31  & 147.68  & 169.08 & 173.58   \\ \toprule[1pt]\toprule[1pt]
\end{tabular}}
\label{TABLE-E}
\end{table*}

\subsubsection{Effectiveness of ARD}
\begin{figure}[t]
	\centering\includegraphics[width=0.47\textwidth,height=4.2cm]{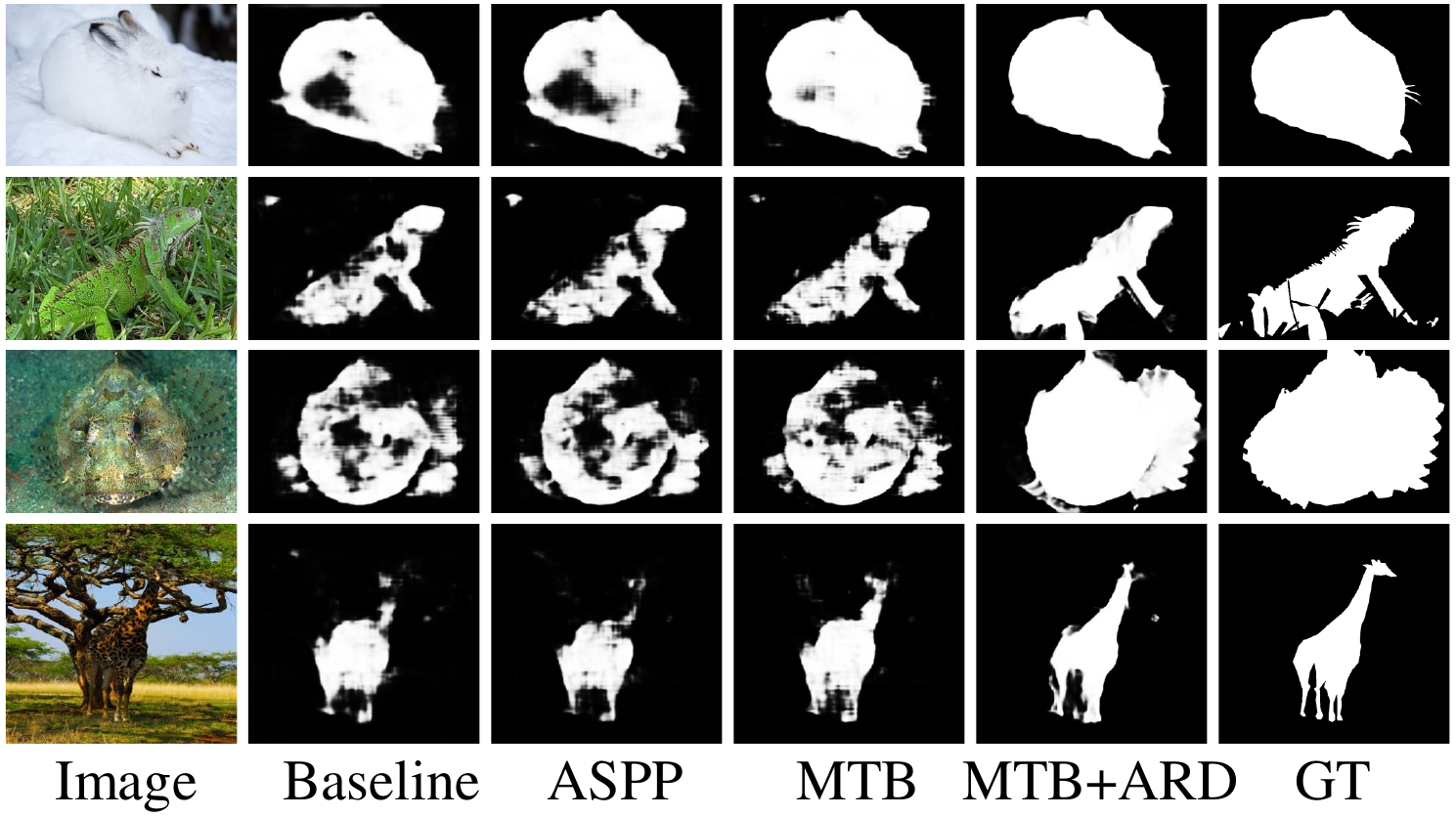}
	\captionsetup{font={small}, justification=raggedright}
	\caption{ Visual comparisons between our modules. }
	\label{Fig.10}
\end{figure}

Adjacent reverse decoder aims to alleviate the problem of gradual semantic dilution of traditional feature pyramid networks, which can integrate diverse information from different levels through cross-layer aggregation and reverse optimization to sufficiently decode significant information. From Table \ref{TABLE-A-1}, when incorporated into ADR ($7^{th}$ row), the performance of the model is further enhanced. Furthermore, we independently verify the effectiveness of MTB and ARD, the global information captured through the MTB ($6^{nd}$ row) is conducive to inferring camouflaged objects than the baseline ($1^{st}$ row). Furthermore, it can be seen that our ARD ($7^{th}$ and $8^{th}$ rows of Table \ref{TABLE-A-1}) is effective for the optimization of input features. In addition, in Fig. \ref{Fig.10}, we give a visual comparison. It can be seen that our ARD facilitates the segmentation of camouflaged objects, which demonstrates the effectiveness of our propose cross-layer aggregation and reverse optimization in the ADR.
\begin{table*}[]
	\renewcommand{\arraystretch}{0.9}
	\setlength{\tabcolsep}{3pt}
	\caption{ Quantitative results on four polyp segmentation datasets. The best two results are shown in \textbf{\color{red} red} and \textbf{\color{green} green}. The symbols ``$\uparrow$/$\downarrow$'' denote that a higher/lower value is better.}
	\centering
	\resizebox*{\textwidth}{27mm}{
		\begin{tabular}{c|ccccc|ccccc|ccccc|ccccc}
			\toprule[1pt]\toprule[1pt]
			\multirow{2}{*}{\normalsize Method} & \multicolumn{5}{c|}{CVC-300 (60 images)}      & \multicolumn{5}{c|}{ETIS (196 images)}       & \multicolumn{5}{c|}{Kvasir (100 images)}     & \multicolumn{5}{c}{CVC-ColonDB (380 images)} \\
			& $MAE$ $\downarrow$   & $AF_m$ $\uparrow$   & $WF_m$  $\uparrow$   & $S_m$ $\uparrow$    & \multicolumn{1}{c|}{$E_m$ $\uparrow$}    & $MAE$ $\downarrow$   & $AF_m$ $\uparrow$   & $WF_m$ $\uparrow$   & $S_m$ $\uparrow$    & \multicolumn{1}{c|}{$E_m$ $\uparrow$}     & $MAE$ $\downarrow$   & $AF_m$ $\uparrow$   & $WF_m$ $\uparrow$   & $S_m$ $\uparrow$    & \multicolumn{1}{c|}{$E_m$ $\uparrow$}    & $MAE$ $\downarrow$   & $AF_m$ $\uparrow$   & $WF_m$  $\uparrow$   & $S_m$ $\uparrow$    & $E_m$ $\uparrow$    \\ \hline
			\rowcolor{blue!8}UNet \cite{UNet}                    & 0.022 & 0.703 & 0.684 & 0.842 & 0.867 & 0.036 & 0.394 & 0.366 & 0.682 & 0.645 & 0.055 & 0.832 & 0.794 & 0.858 & 0.901 & 0.059 & 0.562 & 0.491 & 0.710 & 0.758 \\ 
			UNet++ \cite{UNet++}                  & 0.018 & 0.706 & 0.687 & 0.838 & 0.884 & 0.035 & 0.465 & 0.390 & 0.681 & 0.704 & 0.048 & 0.853 & 0.808 & 0.862 & 0.907 & 0.061 & 0.555 & 0.467 & 0.692 & 0.759 \\ 
			\rowcolor{blue!8}SFA \cite{SFA}                   & 0.065 & 0.353 & 0.341 & 0.640 & 0.604 & 0.109 & 0.255 & 0.231 & 0.557 & 0.515 & 0.075 & 0.715 & 0.670 & 0.782 & 0.828 & 0.094 & 0.393 & 0.366 & 0.629 & 0.634 \\ 
			ACSNet \cite{ACSNet}                 & 0.127 & 0.781 & 0.825 & \textbf{\color{red}0.922} & 0.916 & 0.059 & 0.536 & 0.530 & 0.750 & 0.774 & 0.032 & 0.903 & 0.882 & \textbf{\color{green}0.920} & 0.944 & \textbf{\color{green}0.039} & 0.703 & 0.697 & 0.829 & 0.860 \\ 
			\rowcolor{blue!8}EUNet \cite{EUNet}                  & 0.015 & 0.792 & 0.805 & 0.903 & 0.915 & 0.067 & \textbf{\color{green}0.639} & \textbf{\color{green}0.636} & \textbf{\color{green}0.788} & \textbf{\color{green}0.806} & \textbf{\color{green}0.028} & \textbf{\color{green}0.900} & \textbf{\color{green}0.893} & 0.916 & \textbf{\color{green}0.945} & 0.045 & \textbf{\color{green}0.737} & \textbf{\color{green}0.730} & \textbf{\color{green}0.831} & \textbf{\color{green}0.867} \\ 
			Ours-R                  & \textbf{\color{green}0.009} & \textbf{\color{green}0.831} & \textbf{\color{green}0.830} & \textbf{\color{green} 0.911} & \textbf{\color{green}0.937} & \textbf{\color{green}0.030} & 0.573 & 0.533 & 0.754 & 0.801 & 0.040 & 0.842 & 0.823 & 0.872 & 0.924 & 0.041 & 0.696 & 0.667 & 0.801 & 0.858 \\ 
			\rowcolor{blue!8}Ours-S                  & \textbf{\color{red}0.008} & \textbf{\color{red}0.840} & \textbf{\color{red}0.849} & \textbf{\color{red}0.922} & \textbf{\color{red}0.952} & \textbf{\color{red}0.016} & \textbf{\color{red}0.707} & \textbf{\color{red}0.691} & \textbf{\color{red}0.837} & \textbf{\color{red}0.879} & \textbf{\color{red}0.021} & \textbf{\color{red}0.916} & \textbf{\color{red}0.903} & \textbf{\color{red}0.923} & \textbf{\color{red}0.956} & \textbf{\color{red}0.030} & \textbf{\color{red}0.766} & \textbf{\color{red}0.750} & \textbf{\color{red}0.845} & \textbf{\color{red}0.891} \\ \toprule[1pt]\toprule[1pt]
		\end{tabular}}
        \label{TABLE-PS}
	\end{table*}

\subsubsection{Hyper-parameter experiments}
We conduct hyper-parameter analysis for input image sizes and overall channels. Specifically, we first set overall channels to a fixed 128 for the validation input image sizes, or fix input image sizes in 384$\times$384 to validate overall channels. From Table \ref{TABLE-A-H}, it can be seen that the proposed GLCONet method shows great predicted performance at different overall channels or input image sizes. As the number of overall channels and input image sizes increases, the performance obviously improves, which demonstrates the effectiveness of our GLCONet method. However, the increasing of overall channels or input image sizes will slow down the inference of the network and make the model fitting slow, so we set the input size to 384$\times$384 and the channel to 128 in our GLCONet.

 \begin{figure}[t]
	\centering\includegraphics[width=0.48\textwidth,height=3.3cm]{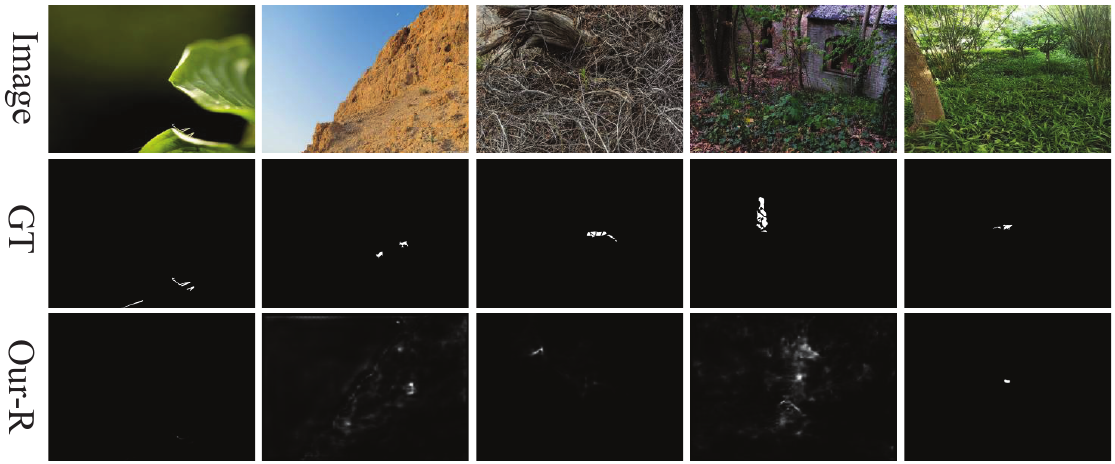}
	\captionsetup{font={small}, justification=raggedright}
	\caption{ Failure results in some very complex scenarios. }
	\label{Fig.11}
\end{figure}

\subsection{Efficiency Analysis and Limitations}
Table \ref{TABLE-E} provides the parameters and FLOPs of our GLCONet model and other existing methods. It can be seen that our method has certain shortcomings in terms of parameter and inference speed. One of the main reasons is that when feature mapping to multi-scale spaces, the number of features to be optimized is large, resulting in large parameters and slower inference speed. Nonetheless, compared to methods JSOCOD \cite{JSOCOD} and FSPNet \cite{FSPNet}, the proposed GLCONet method still has certain advantages. In future work, we will focus on more efficient optimization strategies to explore models that balance efficiency and performance. Additionally, as depicted in Fig. \ref{Fig.11}, when camouflaged targets in the scene are extremely small and heavily obscured, our GLCONet method faces difficulties in making precise predictions for these objects. To address this issue, we propose two potential strategies for future research: \textbf{1)} Enhance the feature representation of the RGB image by incorporating information from different modalities. Specifically, we can strength the segmentation of camouflaged objects by incorporating infrared images, as their imaging principles rely on objects emitting or reflecting infrared radiation. This makes them less affected by camouflage objects and their surroundings. Furthermore, another potential multi-modal information may be text. The rapid development of visual language models offers substantial technical support for object segmentation, aiding in the identification of highly challenging objects by integrating textual information. \textbf{2)} Design a loss function specifically tailored to small or occluded objects, which can enhance the model's ability to predict such objects by increasing its robustness to small and occluded objects. These are merely our viewpoints, and their feasibility will need to be verified in our future works.
\subsection{Expanded Applications}
Automated polyp segmentation can reduce the missed rate of polyps in the colon by medical personnel, which is important for the diagnosis and treatment of colon cancer. Similar to camouflaged object detection, the difficulty of polyp segmentation is that the polyp has only slight differences from its surroundings. To further demonstrate the effectiveness of our GLCONet method, we extend the GLCONet model to the polyp segmentation task.
\subsubsection{Experimental details}
The parameter settings in this paper remain unchanged. We evaluate the GLCONet method on four public polyp segmentation datasets, including CVC-300 \cite{CVC300}, ETIS \cite{ETIS}, Kvasir \cite{Kvasir}, and CVC-ColonDB \cite{cvc}. We exploit the same 900 and 548 images from ClinicDB \cite{cdb} and Kvasir \cite{Kvasir} datasets to train the GLCONet model. We use five evaluation metrics in the COD task to test the performance of the GLCONet model in the polyp segmentation task, involving mean absolute error ($MAE$), average F-measure ($AF_m$), weight F-measure ($WF_m$), S-measure ($S_m$) \cite{Sm}, and E-measure ($E_m$). 

\subsubsection{Result comparisons}
\begin{figure}[]
	\centering\includegraphics[width=0.48\textwidth,height=6.8cm]{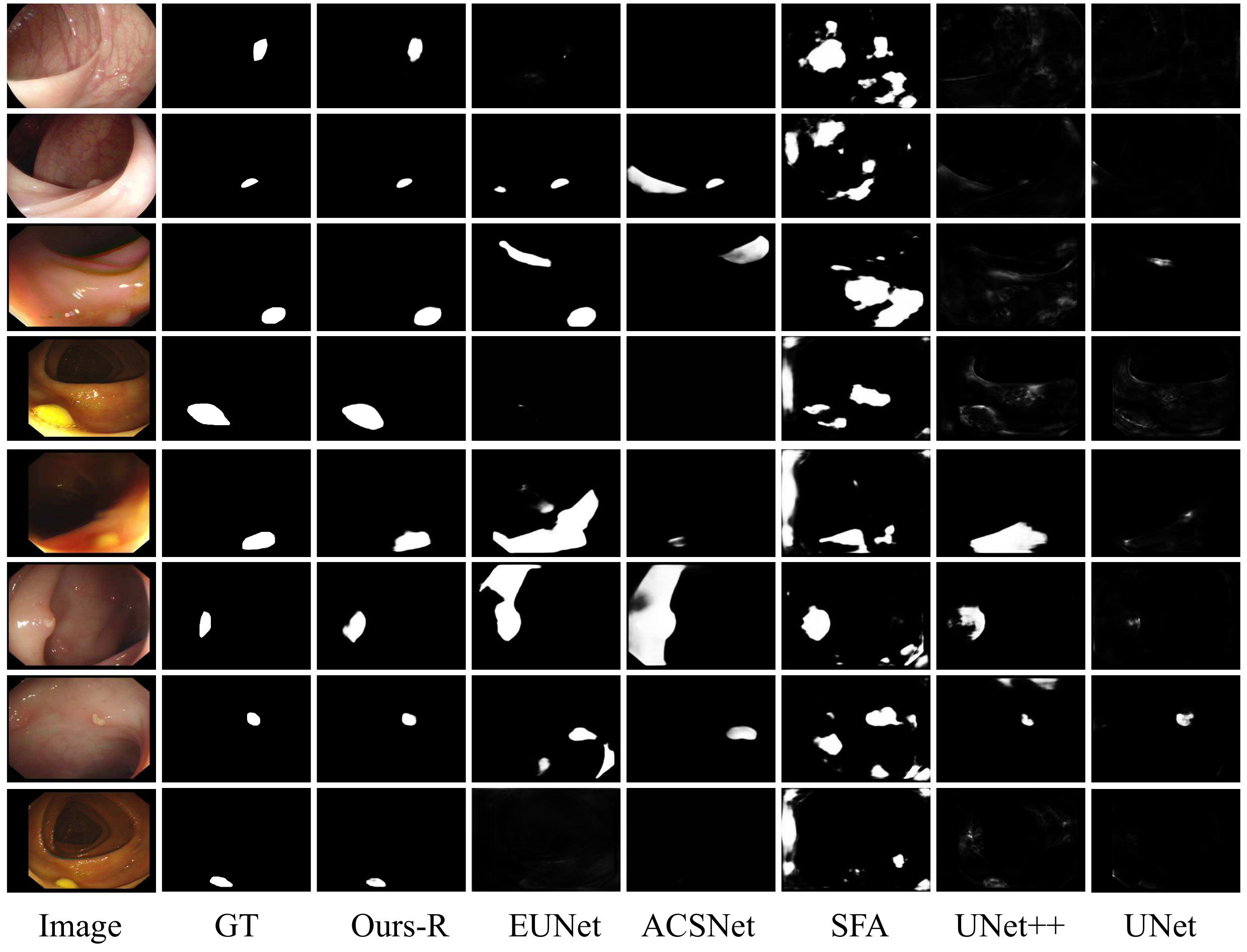}
	\captionsetup{font={small}, justification=raggedright}
	\caption{Qualitative visual comparisons of our GLCONet method and existing polyp segmentation methods.}
	\label{Fig.12}		
\end{figure}

To validate the effectiveness of our GLCONet in the polyp segmentation task, we compare the  GLCONet method with five SOTA methods, including UNet \cite{UNet}, UNet++ \cite{UNet++}, SFA \cite{SFA}, ACSNet \cite{ACSNet}, and EUNet \cite{EUNet}. From Table \ref{TABLE-PS}, it can be seen that our GLCONet method has achieved significant advantages compared to existing methods. In addition, we show visual comparison results, as depicted in Fig. \ref{Fig.12}, our GLCONet method can accurately segment polyps from input images. These above results demonstrate that our GLCONet model has a strong generalization ability. This implies that using the collaborative optimization strategy (COS) and the adjacent reverse decoder (ARD) is beneficial for inferring objects with subtle differences from backgrounds.

\section{Conclusions}
In this paper, we propose a novel global-local collaborative optimization network for camouflaged object detection, dubbed GLCONet. The key to our GLCONet is the collaborative optimization strategy (COS) and the adjacent reverse decoder (ARD). Firstly, in the COS, we develop a global perception module that models long-range dependencies by utilizing multi-scale transformer blocks and exploits a local refinement module that captures local spatial details by using progressive convolution blocks, in which multi-source information is integrated into the group-wise hybrid interaction module. Furthermore, we design an adjacent reverse decoder that contains cross-layer aggregation and reverse optimization to gradually decode the diverse information from multiple-layer features. Extensive experiments on three widely-used COD datasets demonstrate the superiority of our GLCONet method compared to twenty state-of-the-art (SOTA) COD methods under different evaluation metrics.

\ifCLASSOPTIONcaptionsoff
\newpage
\fi

\bibliographystyle{./IEEEtran}
\bibliography{./GLCONet}
\begin{IEEEbiography}
[{\includegraphics[width=1in,height=1.25in,clip,keepaspectratio]{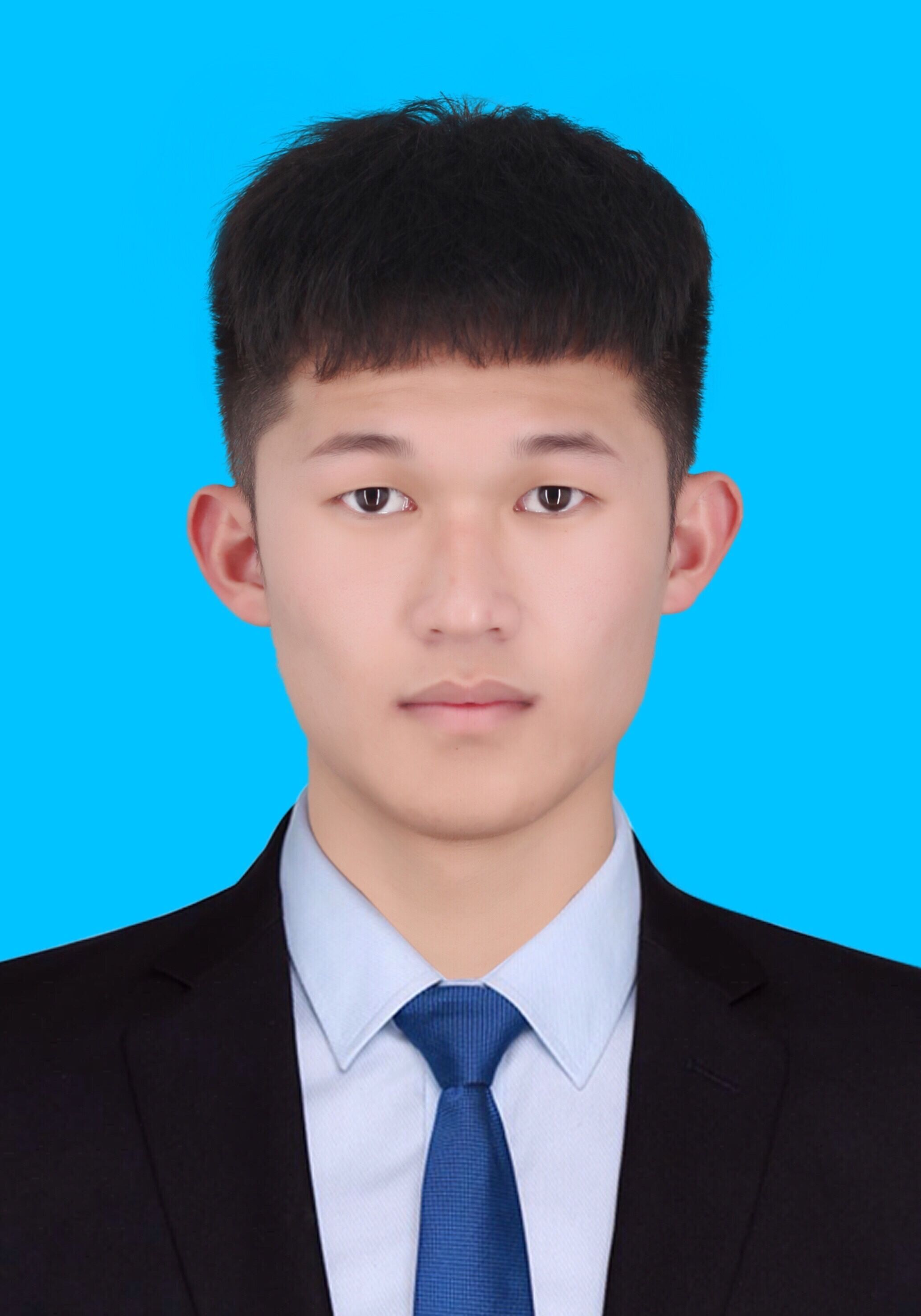}}]{Yanguang Sun} received the M.S. degree from Anhui University of Science and Technology (AUST), Huainan, Anhui, China, in 2023, majoring in software engineering. He is currently pursuing his Ph.D. at Nanjing University of Science and Technology (NJUST), Nanjing, Jiangsu, China, under the supervision of Professor Lei Luo. His research interests focus on computer vision and image analysis. He has served as a reviewer for prestigious conferences such as CVPR and AAAI.	
\end{IEEEbiography}

\begin{IEEEbiography}
[{\includegraphics[width=1in,height=1.25in,clip,keepaspectratio]{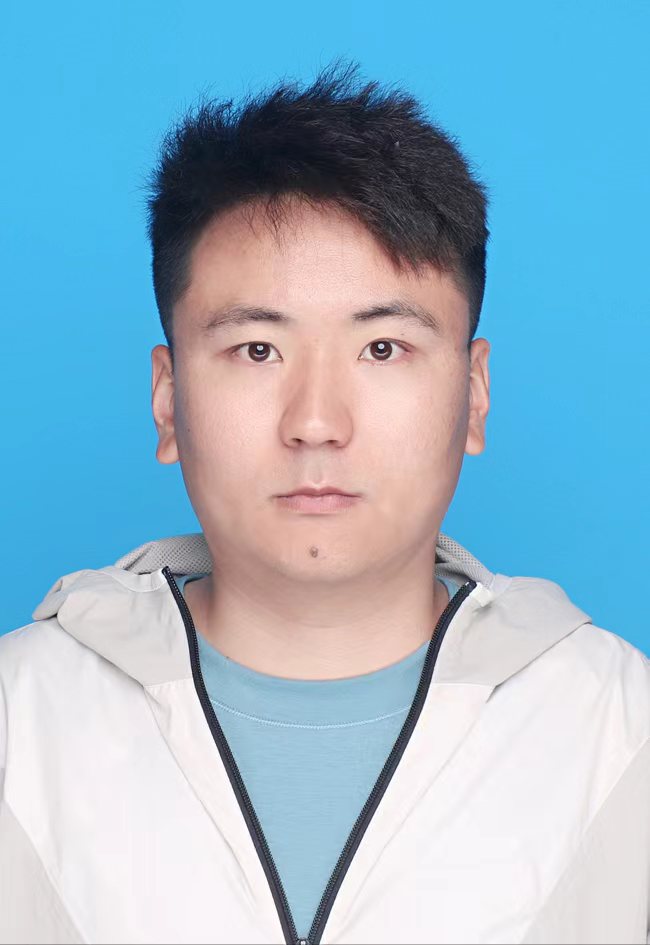}}]{Hanyu Xuan} received the PhD degree from Nanjing University of Science and Technology (NJUST) in 2022, majoring in computer science. He is currently a Lecturer in the School of Big Data and Statistics of Anhui Universily. His research interests include multi-modal learning and computer vision. He has served as a reviewer for over five international conference, such as CVPR, AAAI and ECCV.	
\end{IEEEbiography}

\begin{IEEEbiography}
[{\includegraphics[width=1in,height=1.25in,clip,keepaspectratio]{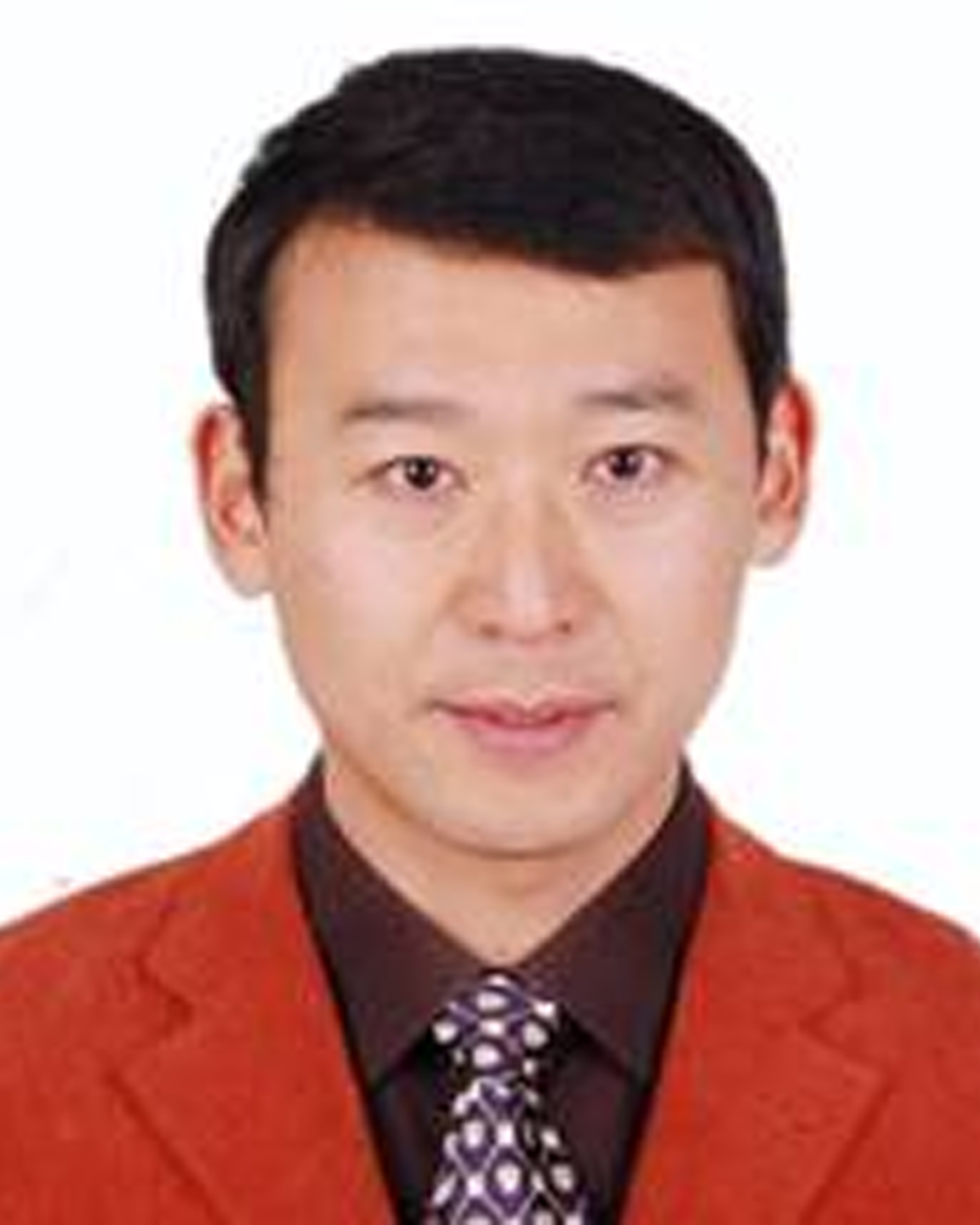}}]{Jian Yang} received the PhD degree from Nanjing University of Science and Technology (NJUST) in 2002, majoring in pattern recognition and intelligence systems. From 2003 to 2007, he was a Postdoctoral Fellow at the University of Zaragoza, Hong Kong Polytechnic University and New Jersey Institute of Technology, respectively. From 2007 to present, he is a professor in the School of Computer Science and Technology of NJUST. He is the author of more than 300 scientific papers in pattern recognition and computer vision. His papers have been cited over 40000 times in the Scholar Google. His research interests include pattern recognition and computer vision. Currently, he is/was an associate editor of Pattern Recognition, Pattern Recognition Letters, IEEE Trans. Neural Networks and Learning Systems, and Neurocomputing. He is a Fellow of IAPR.	
\end{IEEEbiography}

\begin{IEEEbiography}
[{\includegraphics[width=1in,height=1.25in,clip,keepaspectratio]{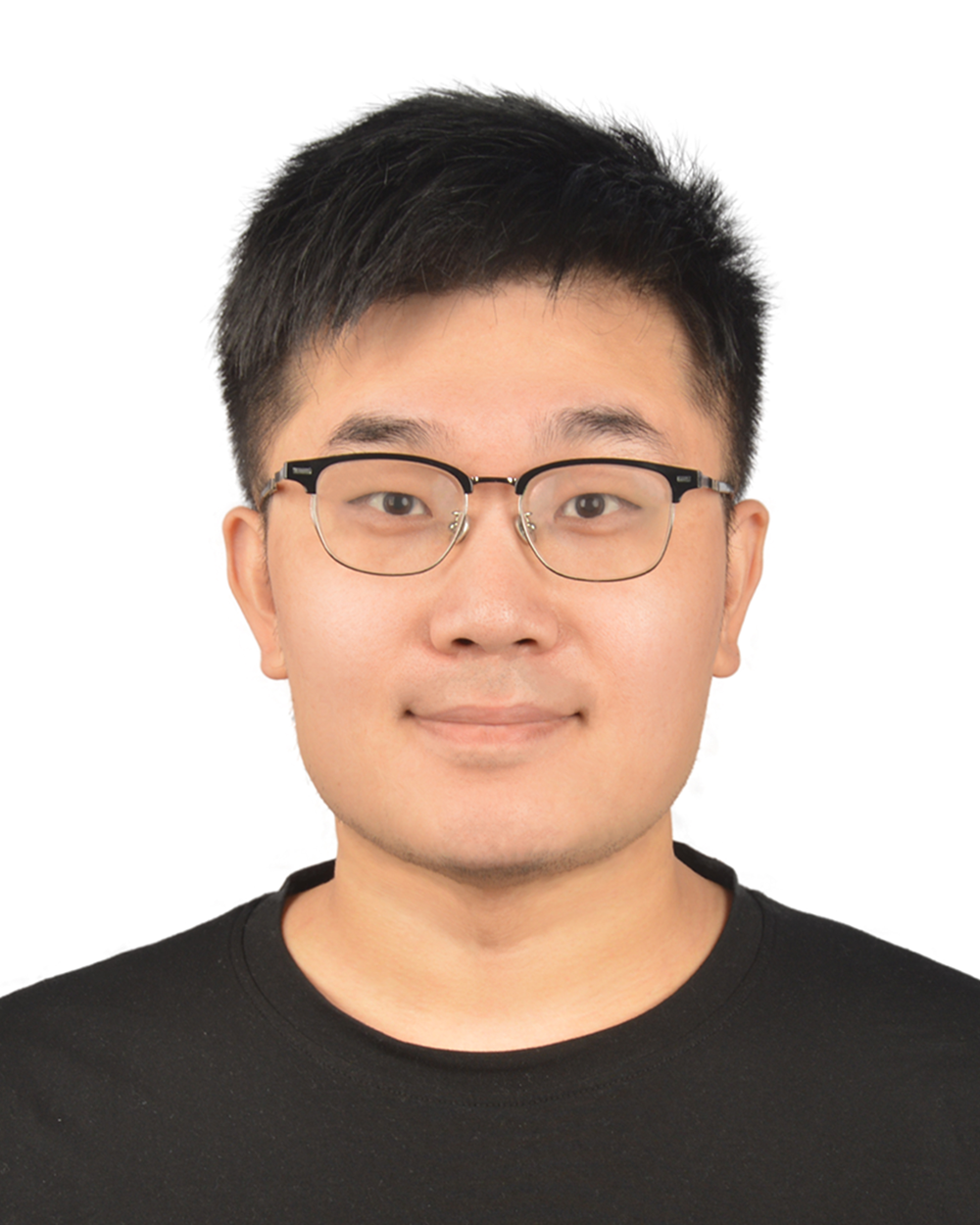}}]{Lei Luo} received the Ph.D. degree in pattern recognition and intelligence systems from the School of Computer Science and Engineering, Nanjing University of Science and Technology (NJUST), Nanjing, China. From 2017 to 2020, he was a Post-Doctoral Fellow at the University of Texas at Arlington, TX, USA, and the University of Pittsburgh, PA, USA. He is currently a Professor in the School of Computer Science and Technology of NJUST. His research interests include pattern recognition, machine learning, data mining and computer vision. Prof. Luo has served as an PC/SPC Member for IJCAI, AAAI, NeurIPS, ICML, KDD, CVPR, and ECCV, and a reviewer for over ten international journals, such as IEEE TPAMI, IEEE TIP, IEEE TSP, IEEE TCSVT, IEEE TNNLS, IEEE TKDE, and PR.	
\end{IEEEbiography}

\end{document}